\documentclass[a4paper]{article}

\usepackage[utf8x]{inputenc}
\usepackage[greek,english]{babel}
\usepackage[T1]{fontenc}
\usepackage{array}
\usepackage{makecell}
\usepackage{tabularx}
\usepackage{tabulary}

\usepackage[a4paper,top=3cm,bottom=2cm,left=3cm,right=3cm,marginparwidth=1.75cm]{geometry}

\usepackage{amsmath}
\usepackage{lscape}
\usepackage{graphicx}
\usepackage[colorinlistoftodos]{todonotes}
\usepackage[colorlinks=true, allcolors=blue]{hyperref}

\title{Text Segmentation using Named Entity Recognition and Co-reference Resolution in English and Greek Texts}
\author{PAVLINA FRAGKOU}

\begin{document}
\maketitle

\begin{abstract}
In this paper we examine the benefit of performing named entity recognition (NER) and co-reference resolution to an English and a Greek corpus used for text segmentation. The aim here is to examine whether the combination of text segmentation and information extraction can be beneficial for the identification of the various topics that appear in a document. NER was performed manually in the English corpus and was compared with the output produced by publicly available annotation tools while, an already existing tool was used for the Greek corpus. Produced annotations from both corpora were manually corrected and enriched to cover four types of named entities. Co-reference resolution i.e., substitution of every reference of the same instance with the same named entity identifier was subsequently performed. The evaluation, using five text segmentation algorithms for the English corpus and four for the Greek corpus leads to the conclusion that, the benefit highly depends on the segment's topic, the number of named entity instances appearing in it, as well as the segment's length.
\end{abstract}

\section{Introduction}
The information explosion of the web aggravates the problem of effective information retrieval. To address this, various techniques, such as text segmentation and information extraction, provide partial solutions to the problem. Text segmentation methods are useful in identifying the different topics that appear in a document. The goal of text segmentation is to divide a text into homogeneous segments, so that each segment corresponds to a particular subject, while contiguous segments correspond to different subjects. In this manner, documents relevant to a query can be retrieved from a large database of unformatted (or loosely formatted) texts. Two main versions of the text segmentation problem appear in the literature. The first version concerns segmentation of a single large text into its constituent parts (e.g., to segment an article into sections). The second version concerns segmentation of a stream of independent, concatenated texts (e.g., to segment a transcript of a TV news program into separate stories). Text segmentation proves to be beneficial in a number of scientific areas, such as corpus linguistics, discourse segmentation, and as a preliminary step to text summarization.

Information extraction is the task of automatically extracting structured information from unstructured and/or semi-structured documents. In most of the cases, this activity  concerns processing texts by means of automated natural language processing (NLP) tools. Information extraction methods try to identify portions of text that refer to a specific topic, by focusing on the appearance of instances of specific types of named entities (such as person, organization, date, and location) according to the thematic area of interest.

The question that arises is whether the combination of text segmentation and information extraction (and more specifically the Named Entity Recognition (NER) and co-reference resolution steps) can be beneficial for the identification of the various topics appearing in a document. In other words, the exploitation by a text segmentation algorithm of the information that is provided by named entity instances given that, the former examines the distribution of words appearing in a document but not the content of the selected words, i.e. it does not exploit the importance that several words may have in a specific context (such as person names, locations, dates etc).

This paper examines the benefit of performing NER and co-reference resolution in corpora belonging to two different languages i.e., English and Greek. More specifically for English, Choi's corpus (\cite{Choi:2000}) which is used as benchmark for examining the performance of text segmentation algorithms was examined. For Greek, the corpus presented in Fragkou, Petridis and Kehagias (\cite{Fragkou:2007}) - which was previously applied to three text segmentation algorithms - consisting of portions of texts taken from the Greek newspaper 'To Vima' was considered for examination.

We stress that the focus is not on finding the algorithm that achieves the best segmentation performance on the corpora, rather on the benefit of performing NER and co-reference resolution on a corpus used for text segmentation.

The structure of the paper is as follows. Section 2 provides an overview of related methods. Section 3 presents the steps performed for the creation of the 'annotated' corpus for English, while Section 4 presents the same steps for the Greek corpus. Section 5 provides a description of the text segmentation algorithms chosen for our experiments. Section 6 presents the evaluation metrics used. Section 7 lists evaluation results obtained using five (four) well known text segmentation algorithms to the English and the Greek corpus (respectively), while Section 8 provides conclusions and future steps.

\section{Related Work}

According to Wikipedia, topic analysis consists of two main tasks: topic identification and text segmentation. While the first is a simple classification of a specific text, the latter case implies that a document may contain multiple topics, thus the task of computerized text segmentation may be to discover these topics automatically and segment the text accordingly, by dividing it into meaningful units. Automatic segmentation is the problem of implementing a computer process to segment text i.e., \emph{'given a text which consists of several parts (each part corresponding to a different subject - topic), it is required to find the boundaries between the parts'}. When punctuation and similar clues are not consistently available, the segmentation task often requires techniques, such as statistical decision-making, large dictionaries, as well as consideration of syntactic and semantic constraints. Effective natural language processing systems and text segmentation tools usually operate on text in specific domains and sources.

A starting point is the calculation of the within-segment similarity based on the assumption that parts of a text having similar vocabulary are likely to belong to a coherent topic segment. It is worth noticing that within-segment similarity is calculated on the basis of words, but not on the basis of the application of other more sophisticated techniques such as NER or co-reference resolution which in their turn, highlight the appearance of specific words in the scope of a particular topic. In the literature, several word co-occurrence statistics are proposed (\cite{{Choi:2000},{Choi:2001},{Hearst:1997},{Utiyama:2001}}). 

A significant difference between text segmentation methods is that some evaluate the similarity between all parts of a text (\cite{{Choi:2000},{Choi:2001}}), while others between adjacent parts (\cite{{Hearst:1997},{Utiyama:2001}}). To penalize deviations from the expected segment length, several methods use the notion of 'length model' (\cite{Heinonen:1998}). Dynamic programming is often used to calculate the globally minimal segmentation cost (\cite{{Xiang:2003},{Heinonen:1998},{Kehagias:2004},{Qi:2008}}).

Other approaches involve the improvement of the dotplotting technique (\cite{Ye:2005}), the improvement of Latent Semantic Analysis (\cite{Bestgen:2006}), and the improvement of the TextTiling method (\cite{Hearst:1997}) presented by method  Kern and Granitzer (\cite{Kern:2009}). Recent work in text segmentation, involves among others Affinity Propagation to create segment centers and segment assignment for each sentence (\cite{Kazantseva:2011}), the Markovian assumption along with Utiyama and Isahara' s algorithm (\cite{Simon:2013}) as well as unsupervised lecture segmentation (\cite{Malioutov:2006}). Yu et al. (\cite{Yu:2012}) propose a different approach in which, each segment unit is represented by a distribution of the topics, instead of a set of word tokens thus, a text input is modeled as a sequence of segment units and Markov Chain Monte Carlo technique is employed to decide the appropriate boundaries.

While several papers examine the problem of segmenting English texts, little work is performed for Greek. An important difference between segmenting English and Greek texts lies in the higher degree of inflection that Greek language presents. This makes the segmentation problem even harder. To the author's best knowledge, the only work that refers to segmentation of Greek texts appears in \cite{Fragkou:2007}.

On the other hand, information extraction aims to locate inside a text passage domain-specific and pre-specified facts (e.g., in a passage about athletics, facts about the athlete participating in a 100m event, such as his name, nationality, performance, as well as facts about the specific event, such as its name). Information extraction (IE) can be defined as the automatic identification of selected types of entities, relations or events in free text (\cite{Grishman:1997}).

Two of the fundamental processing steps usually followed to find the aforementioned types of information, are (\cite{Appelt:1993}): (a)\emph{Named Entity Recognition (NER)}, where entity mentions are recognized and classified into proper types for the thematic domain such as persons, places, organizations, dates , etc.; (b) Co-reference, where all mentions that represent the same entity are identified and grouped together according to the entity they refer to, such as \emph{'Tatiana Lebedeva', 'T. Lebedeva'}, or \emph{'Lebedeva'}.

Named entity annotation for English is extensively examined in the literature and a number of automated annotation tools exist. The majority of them such as GATE (\cite{Cunningham:2011}), Stanford NLP tools, ((\cite{Lee:2013}), \url{http://nlp.stanford.edu/index.shtml}), Illinois NLP tools (\url{http://cogcomp.cs.illinois.edu/page/tools/}), Apache OpenNLP library (\url{http://opennlp.apache.org/}), LingPipe (\url{http://alias-i.com/lingpipe/}) contain a number of reusable text processing toolkits for various computational problems such as tokenization, sentence segmentation, part of speech tagging, named entity extraction, chunking, parsing, and frequently, co-reference resolution. GATE additionally comes with numerous reusable text processing components for many natural languages.

Another category of NER tools involves stand - alone automated linguistic annotation tools such as Callisto for task-specific annotation interfaces e.g., named entities, relations, time expressions etc (\cite{Day:1997}), MMAX2 which uses stand-off XML and annotation schemas for customization (\cite{Muller:2006}, and Knowtator which supports semi-automatic adjudication and the creation of a consensus annotation set (\cite{Ogren:2006}).

NER task is limited dealt for Greek texts. Work on Greek NER usually relies on hand-crafted rules or patterns (\cite{{Boutsis:2000}, {Farmakiotou:2000}, {Farmakiotou:2002}}) and/or decision tree induction with C4.5 (\cite{{Karkaletsis:1999}, {Petasis:2001}}). Diamantaras, Michailidis and Vasileiadis (\cite{Diamantaras:2005}), and Michailidis et al. (\cite{Michailidis:2006}) are the only exceptions, where SVMs, Maximum Entropy, Onetime, and manually crafted post-editing rules were employed. Two works deserve special attention. The first examines the problem of pronominal anaphora resolution (\cite{Papageorgiou:2002}). The authors created an Information extraction pipeline which included a tokenizer, a POS tagger, and a lemmatizer, with tools that recognize named entities, recursive syntactic structures, grammatical relations, and co-referential links.

The second work is the one proposed by Lucarelli et al. (\cite{Lucarelli:2007}), where a freely available named entity recognizer for Greek texts was constructed . The recognizer identifies temporal expressions, person names, and organization names. Another novelty of this system is the use of active learning, which allows it to select by itself candidate training instances to be annotated by a human during training.

Co-reference resolution includes -among others- the step of anaphora resolution. The term 'anaphora' denotes the phenomenon of referring to an entity already mentioned in a text, most often with the help of a pronoun or a different name. Co-reference involves basically the following steps: (a) pronominal co-reference• finding the proper antecedent for personal pronouns, possessive adjectives, possessive pronouns, reflexive pronouns, pronouns 'this' and 'that'; (b) identification of cases where both the anaphor and the antecedent refer to identical sets or types. This identification requires some world knowledge or specific knowledge of the domain. It also includes cases such as reference to synonyms or cases where the anaphor matches exactly or is a substring of the antecedent; (c) ordinal anaphora (for cardinal numbers and adjectives such as 'former' and 'latter').

Some well-known tools performing co-reference resolution are Reconcile (\cite{Stoyanov:2010}), BART (\cite{Versley:2008}), Illinois Co-reference Package (\cite{Bengtson:2008}), and Guitar (General Tool for Anaphora Resolution). The majority of those tools utilize supervised machine learning classifiers taken for example from the Weka tool (i.e., Reconcile and BART) as well as other language processing tools. Reconcile additionally offers the ability to run on unlabeled texts  (\cite{Stoyanov:2010}). Illinois Co- reference Package contains a co-reference resolver along with co-reference related features including gender and number match, WordNet relations including synonym, hypernym, and antonym, and finally ACE entity types (\cite{Bengtson:2008}). Other tools focus on specific co-reference tasks such as Guitar (General Tool for Anaphora Resolution, \url{http://cswww.essex.ac.uk/Research/nle/GuiTAR/}) which focuses on anaphora resolution.

Co-reference resolution was also applied as a subsequent step of NER for Greek. More specifically, Papageorgiou et al. chose to focus on pronominal anaphora resolution i.e., the task of resolving anaphors that have definite description as their antecedents among the broad set of referential phenomena that characterize the Greek language (\cite{Papageorgiou:2002}). The pronoun types that were selected for annotation were the third person possessive and the relative pronoun. Two forms of anaphora were covered: (a) intra-sentential, where co-referring expressions occur in the same sentence and (b) inter-sentential, where the pronoun refers to an entity mentioned in a previous sentence.

The combination of paragraph or discourse segmentation with co-reference resolution presents strong similarity to the segmentation of concatenated texts. Litman and Passonneau (\cite{{Litman:1995a}, {Litman:1995b}}) use a decision tree learner for segmenting transcripts of oral narrative texts using three sets of cues: prosodic cues, cue phrases, and noun phrases (e.g., the presence or absence of anaphora). Barzilay and Lapata (\cite{{Barzilay:2005a}, {Barzilay:2005b}, {Barzilay:2008}}) presented an algorithm which tries to capture semantic relatedness among text entities by defining a probabilistic model over entity transition sequences distribution. Their results validate the importance of the combination of co-reference, syntax, and salience. In their corpus, the benefit of full co-reference resolution is less uniform due to the nature of the documents. In Singh et al. (\cite{Singh:2013}), the authors propose a single joint probabilistic graphical model for classification of entity mentions (entity tagging), clustering of mentions that refer to the same entity (co-reference resolution), and identification of the relations between these entities (relation extraction). Special interest presents the work conducted in Yao et al. (\cite{Yao:2013}), where the authors present an approach to fine-grained entity type classification by adopting the universal schema whose key characteristic is that it models directed implicature among the many candidate types of an entity and its co-reference mentions.

The importance of text segmentation and information extraction is apparent in a number of applications, such as noun phrase chunking, tutorial dialogue segmentation, social media segmentation such as Twitter or Facebook posts, text summarization, semantic segmentation, web content mining, information retrieval, speech recognition, and focused crawling. The potential use of text segmentation in the information extraction process was examined in Fragkou (\cite{Fragkou:2009}). Here, the reverse problem is examined i.e., the use of information extraction techniques in the text segmentation process. Those techniques are applied on two different corpora used for text segmentation, resulting in the creation of two 'annotated' corpora. Existing algorithms performing text segmentation exploit a variety of word co-occurrence statistic techniques in order to calculate the homogeneity between segments, where each segment refers to a single topic. However, they do not exploit the importance that several words may have in a specific context. Examples of such words are person names, locations, dates, group of names, and scientific terms. The importance of those terms is further diminished by the application of word pre-processing techniques i.e., stop-list removal and stemming on words such as pronouns. More specifically, all types of pronouns for English which have proved to be useful for co-reference resolution are included in the stop list used in Information Retrieval area (a snapshot of the list can be found in \url{http://jmlr.org/papers/volume5/lewis04a/a11-smart-stop-list/english.stop} or in \url{http://www.textfixer.com/resources/common-english-words.txt}). This however, does not hold for Greek, where publicly available stop word lists do not include any Greek pronouns. The aim of this paper is to exploit whether the identification of such words can be beneficial for the segmentation task. This identification requires the application of NER and co-reference resolution, thus to evaluate the potential benefit resulted from manual effort i.e., annotation or correction, and/ or completion of it in comparison with the application of publicly available automated annotation tools for NER and co-reference resolution.

Consider for example an article that appears in Wikipedia referring to Alan Turing (\url{https://en.wikipedia.org/wiki/Alan\_Turing}). The first paragraph of the article regarding the author is the following: “\textit{Alan Mathison Turing OBE FRS (23 June 1912 - 7 June 1954) was a pioneering \underline{British computer scientist, mathematician,
logician, cryptanalyst and theoretical biologist}. He was highly influential in the development of \underline{theoretical computer science}, providing a formalisation of the concepts of \underline{algorithm} and \underline{computation} with the \underline{Turing machine}, which can be considered a model of a general purpose computer. Turing is widely considered to be the father of theoretical computer science and \underline{artificial intelligence}}.” Underlined words (with the exceptions of: (a) \textit{23 June 1912 - 7 June 1954}, which corresponds to named entity instances of type date; (b) \underline{Alan Mathison Turing} and \underline{Turing}, which correspond to named entity instances of type person; (c) \underline{\textit{British}}, which corresponds to a variation of named entity instance of type Country; and (d) word \underline{He} which corresponds to named entity instance of type Person \underline{Turing}), may be considered as named entity instances of the following types: person, profession, science – scientific terms, etc.

In case where we are posing a query using as keywords the words “\textit{Turing and Enigma}” or “\textit{Turing and cryptography}”, ideally instead of receiving as a result the whole page, we would like to receive portions of texts where expressions marked as named entity instances of a scientific term or a variation of it (such as \textit{Enigma machine, Enigma motors, Enigma code, Enigma-enciphered messages, Enigma signals} etc for the first case and \textit{cryptanalysis, cryptanalyst, cryptology decryption, Cryptography} etc for the second) i.e., Section 3 of the Wikipedia page for both queries. This approach enhances the presence of semantic information -through information extraction techniques- as opposed to treating separately every word. Thus, it (intuitively) reinforces - with the help also of co-reference resolution - the identification of \textbf{portions} of texts that refer to the desired information. Extraction of those portions is performed via text segmentation algorithms.

The aim of this paper is to examine the contribution of semantic information, attributed either manually or using publicly available named entity and co-reference resolution tools, to effectively identifying portion(s) of text corresponding to a specific topic (expressed as a query). In other words, the attempt to add semantic information to parts of a text, and examine the contribution of this information to the identification of desired information. Subsequently, we would like to highlight information that is important to a specific content and how this information is not eliminated by stop list removal and stemming. Attention is paid to group of words that prove to represent important information such as a person or a location which otherwise, this importance is underestimated or eliminated due to word pre-processing. A step further can be to link related information in the perspective of a well-known ontology (linked data).

To the author's best knowledge, a similar work was presented for French in (\cite{Sitbon:2005}). The authors there used two corpora. The first was a manually-built French news corpus, which contained four series of 100 documents. Each document was composed of ten segments extracted from the '\textit{Le Monde}' newspaper. The second corpus focused on a single topic (i.e., sports). In each of those corpora, the authors performed NER using three types of named entities: person name, location, and organization. The authors claim use of anaphors but provide no further details. They used named entity instances as components of lexical chains to perform text segmentation. The authors reported that according to the obtained results on their corpus, use of named entities does not improve segmentation accuracy. The authors state that, an explanation to the obtained results can be the frequent use of anaphora that results in limitation of named entity repetition. Moreover, the use of lexical chains restricts the number of features used.

To the author's best knowledge, no similar work exists for the Greek language in the literature combining NER and co-reference resolution to assist the text segmentation task. It must be stressed that, for both languages, manual annotation including all types of anaphora resolution was performed. The separate contribution of specific types of anaphora (such as pronominal anaphora) in a text segmentation algorithm will be addressed as future work.

Our work improves the one presented in (\cite{Sitbon:2005}) in six points. The first one is the use of a widely accepted benchmark i.e., Choi's text segmentation corpus (\cite{Choi:2000}). Even though reported results from different algorithms are extremely efficient, we have chosen to work on this corpus due to the availability of the aforementioned results and the widely accepted partial annotation of the Brown corpus from which it is constructed, as it is described in details in the next section. The second point is the use of an additional named entity i.e., date\footnote{“Date” as a named entity type is used by the majority of publicly (or not) named entity recognition tools. It was also used in TREC evaluation to question systems regarding who, where and when. In the present work, calculations proved that, 16\% of named entity instances produced by Illinois NER belong to “date” named entity type. The equivalent percentage for manual named entity annotation in Choi’s corpus is lower, approximately 10\%}. The third point is the application of manual co-reference resolution (i.e., all the aforementioned tasks of co-reference resolution) to those portions of text that refer to named entity instances as a subsequent step of NER, after proving that publicly available tools cannot be easily used for the problem in question. The fourth point involves the comparison of manual annotation with the ouput produced by combining publicly available automated annotation tools (i.e. Illinois NER as well as Illinois Co-referencer and Reconcile Co-referencer). The last points involve the evaluation of the produced annotated corpus using five text segmentation algorithms, and the use of an additional high inflectional language i.e., Greek.

\section{English Corpus}

The corpus used here is the one generated by Choi (\cite{Choi:2000}). The description of the 700 samples corpus is as follows: '\textit{A sample is a concatenation of ten text segments. A segment is the first n sentences of a randomly selected document from the Brown Corpus. A sample is characterized by the range n.}' More specifically, Choi’s dataset is divided into four subsets (“3-5”, “6-8”, “9-11” and “3-11”) depending upon the number of sentences in a segment/story. For example, in subset “X-Y”, a segment is derived by (randomly) choosing a story from Brown corpus, followed by selecting first N (a random number between X and Y) sentences from that story. Exactly ten such segments are concatenated to make a document/sample. Further, in each subset there are 100 documents/samples to be segmented, except from subset 3-11 where 400 documents/samples were created. Thus, documents/samples belonging for example in subset 3-5 are not included as is in other subsets i.e., are unique and are listed separately. Each segment is the first n sentences of a randomly selected document from the Brown corpus, s.t. $3\leq n\leq 11$ ≤ for subset 3-11. Table 1 gives the corpus statistics. More specifically, Choi created his corpus by using sentences selected from 44 documents belonging to A category Press and 80 documents belonging to J category Learned. According to Brown Corpus description, category A contains documents about Political, Sports, Society, Sport News, Financial, and Cultural. Category J contains documents about Natural Sciences, Medicine, Mathematics, Social and Behavioral Sciences, Political Science, Law, Education, Humanities, Technology, and Engineering. Documents belonging to category J, usually contain portions of scientific publications about mathematics or chemistry. Thus, they contain scientific terms such as \textit{urethane foam}, \textit{styrenes} and \textit{gyro-stabilized platform system}. On the other hand, the majority of documents of category A usually contain person names, locations, dates, and groups of names.

It must be stressed that, since chosen stories from Brown Corpus are finite (44 stories are included in Category A and 80 in Category J), portions of the same story may appear in many segments in any of the four subsets. Since the creation of each of the ten segments of each of the 700 samples results from \textbf{randomly} selecting a document either from category A or from Category J, there is no rule regarding use of specific documents from category A or J to each of the four subsets. To give an idea of this, the first document of subset 3-5 contains portions of texts belonging to the following documents: J13, J32, A04, J48, J60, J52, J16, J21, J57, J68, where (A or J)XX denotes Category A or J of Brown Corpus and XX the file’s number (among the 44 or 80 of the category).

\begin{center}%
\begin{tabular}
[c]{|r|r|r|r|r|}\hline
\textbf{Range of n } & \textbf{3-11} & \textbf{3-5} & \textbf{6-8}  & \textbf{9-11}\\\hline
No.samples & 400 & 100 & 100 & 100\\\hline
\end{tabular}
\linebreak
\textbf{Table 1}:Choi's Corpus Statistics (Choi 2000).
\end{center}

Recent bibliography in text segmentation involves use of other datasets. Among those is the one compiled by Malioutov and Barzilay (\cite{Malioutov:2006}), which consists of manually transcribed and segmented lectures on Artificial Intelligence. The second dataset consists of 227 chapters from medical textbooks (\cite{Eisenstein:2008}). The third dataset  consists of 85 works of fiction downloaded from Project Gutenberg in which segment boundaries correspond to chapter breaks or to breaks between individual stories. Lastly, the ICSI Meeting corpus is frequently used consisting of 75 word-level transcripts (one transcript file per meeting), time-synchronized to digitized audio recordings.

Even though the aforementioned datasets are used in the literature for the text segmentation task, they are not chosen to be used in the present study. The reason for this is that Choi's dataset used here is strongly related to Semcor which provides a specific type of automated named entity annotation. Morever, to the author's best knowledge, no annotated set appears in the literature for any of the aforementioned datasets.

\subsection{Named Entity Annotation}

There exist a number of readily-available automated annotation tools in the literature. In the work presented by Atdağ and Labatut (\cite{Atdag:2013}), a comparison of four publicly available, well known and free for research NER tools i.e, Stanford NER, Illinois NER, OpenCalais NER WS and Alias-i LingPipe took place in a new corpus created by annotating 247 Wikipedia articles. Atdağ and Labatut claim that: '\textit{NER tools differ in many ways. First, the methods they rely upon range from completely manually specified systems (e.g. grammar rules) to fully automatic machine-learning processes, not to mention hybrids approaches combining both. Second, they do not necessarily handle the same classes of entities. Third, some are generic and can be applied to any type of text, when others focus only on a specific domain such as biomedicine or geography. ... Fifth, the data outputted by NER tools can take various forms, usually programmatic objects for libraries and text files for the others. There is no standard for files containing NER- processed text, so output files can vary a lot from one tool to the other. Sixth, tools reach different levels of performance. Moreover, their accuracy can vary depending on the considered type of entity, class of text, etc. Because of all these differences, comparing existing NER tools in order to identify the more suitable to a specific application is a very difficult task. And it is made even harder by other factors: … in order to perform a reliable assessment, one needs an appropriate corpus of annotated texts. This directly depends on the nature of the application domain, and on the types of entities targeted by the user. It is not always possible to find such a dataset …. Lastly, NER tools differ in the processing method they rely upon, the entity types they can detect, the nature of the text they can handle, and their input/output formats. This makes it difficult for a user to select an appropriate NER tool for a specific situation}'. Moreover, as it is stated in Siefkes (\cite{Siefkes:2007}), '\textit{There are several other assumptions that are generally shared in the field of IE, but are seldom mentioned explicitly. One of them is corpus homogeneity: Since the properties of the relevant extracted information have to be learned from training examples, training corpora should be sufficiently homogeneous, that is the texts in a training corpus are supposed to be similar in expression of relevant information.}'.

The majority of readily - available tools require training, which is usually focused on a single or a limited number of topics. The fact that each tool is trained on a different corpus oblige us to select the one that is trained on a corpus referring to similar topics with the ones appearing in Categories A and J of the Brown Corpus. Additionally, the potential use of existing tools must: a) produce efficient annotation result i.e., no need or  restricted need to perform manual correction (as a result of failure to recognize all named entity types covering all topics mentioned in a text); b) cover all aspects of NER and co- reference resolution; c) attribute a unique named entity identifier to each distinct instance (including all mentions of it); d) produce an output that can be easily be given as input to a text segmentation algorithm.

In order to avoid manual annotation effort and test potential use of readily-available (already trained) automated annotation tools in the literature, we conducted a first trial using publicly available tools. Those tools perform either exclusively co-reference resolution or a number or tasks such as sentence splitting, parsing, part of speech tagging, chuncking, or name entity recognition (with a different predefined number of named entity types). Examination of those tools was performed on a portion of text belonging to Choi's dataset (see Table 2 which lists the output resulting from different tools, paying attention to the type(s) of co-reference that each tool can capture). Due to space limitations, the output of each of the examined tools is restricted to few sentences.

More specifically, the following tools were examined: 
\begin{enumerate}
\item Apache OpenNLP, which provides a co-reference resolution tool (\url{http://opennlp.apache.org/}) 
\item Stanford NER, and more specifically Stanford Named Entity Recognizer \item Illinois NER (\url{http://cogcomp.cs.illinois.edu/page/software_view/NETagger}) and more specifically the latest version of Illinois NER which makes use of 19 named entity types i.e.: $PERSON, $DATE, $ORG, $GPE, $LOC, $CARDINAL, $MONEY, $NORP, $WORK_OF_ART, $EVENT, $TIME, $FAC, $LAW, $ORDINAL, $PERCENT, $QUANTITY, $PRODUCT,$ LANGUAGE, MISC
\item Illinois Co-reference Package, which provides an already trained model for co- reference resolution (\cite{Bengtson:2008})
\item BART - a Beautiful Anaphora Resolution Toolkit which identifies named entity types (such as person and location), as well as noun phrases in addition to co-reference resolution (\cite{Versley:2008}) and 
\item Reconcile - Co-reference Resolution Engine which is the only tool that attributes co- reference mentions to the identified named entity instances (\cite{Stoyanov:2010}).

\end{enumerate}
\pagebreak
\begin{tabulary}{\textwidth}{|L|L|}
\textbf{Original Text} & {\small Vincent G. Ierulli has been appointed temporary assistant district attorney , it was announced Monday by Charles E. Raymond, District Attorney.Ierulli will replace Desmond D. Connall who has been called to active military service but is expected back on the job by March 31.}\\\hline
\textbf{Apache OpenNLP } & {\small [NP ==\_NN] =\_SYM =\_SYM =\_SYM =\_SYM =\_SYM =\_SYM =\_SYM =\_SYM [NP <START:person>\_NNP Vincent\_NNP <END>\_NNP G.\_NNP Ierulli\_NNP] [VP has\_VBZ been\_VBN appointed\_VBN] [NP temporary\_JJ assistant\_NN district\_NN attorney\_NN] ,\_, [NP it\_PRP] [VP was\_VBD announced\_VBN] [PP <START:date>\_IN] [NP Monday\_NNP <END>\_NNP] [PP by\_IN] [NP <START:person>\_NNP Charles\_NNP E.\_NNP Raymond\_NNP <END>\_NNP] ,\_, [NP <START:person>\_NNP District\_NNP Attorney\_NNP <END>\_NNP] .\_.  [NP Ierulli\_NNP] [VP will\_MD replace\_VB] [NP Desmond\_NNP D.\_NNP Connall\_NNP] [NP who\_WP] [VP has\_VBZ been\_VBN called\_VBN] [PP to\_TO] [NP active\_JJ military\_JJ service\_NN] but\_CC [VP is\_VBZ expected\_VBN] [ADVP back\_RB] [PP on\_IN] [NP the\_DT job\_NN] [PP by\_IN] [VP <START:date>\_VBG] [NP March\_NNP 31\_CD <END>\_NNP]}. \\\hline 
\textbf{Stanford NER} & {\small \textbf{Vincent G.Ierulli} has been appointed temporary assistant district attorney, it was announced \textbf{Monday} by \textbf{Charles E. Raymond},\textbf{District Attorney}. Ierulli will replace \textbf{Desmond D. Connall} who has been called to active military service but is expected back on the job by \textbf{March 31}} .\\\hline
\textbf{Illinois NER} & {\small [PERSON Vincent G. Ierulli] has been appointed temporary assistant district attorney , it was announced [DATE Monday] by [PERSON Charles E. Raymond] , District Attorney. \newline [PERSON Ierulli] will replace [PERSON Desmond D. Connall] who has been called to active military service but is expected back on the job by [DATE March 31] }.\\\hline
\textbf{Illinois Co-reference Package } & {**Vincent G. Ierulli*\_8 has been appointed **temporary *assistant district*\_19 attorney*\_21*\_21 , it was announced Monday by *Charles E. Raymond*\_5 , **District Attorney*\_23*\_23.*Ierulli*\_8 will replace *Desmond D. Connall *who*\_16 has been called to active *military*\_15 service*\_16 but is expected back on th job by March 31 }.\\\hline
\textbf{BART - a Beautiful Anaphora Resolution Toolkit } & {\small {person Vincent G. Ierulli} has been {np {np appointed temporary assistant district} attorney}, {np it} was announced {np Monday} by {np {person Charles E. Raymond} , District Attorney}.\newline  {person Ierulli} will replace {person Desmond D. Connall} who has been called to {np active military service} but is expected back on {np the job} by {np March 31} }. \\\hline
\textbf{Reconcile - Co-reference \newline Resolution Engine} & <NP NO="0" CorefID="5">Vincent G. Ierulli</NP> has been appointed temporary <NP NO="1" CorefID="1">assistant district attorney</NP>, <NP NO="2" CorefID="2">it</NP> was announced <NP NO="3" CorefID="3">Monday</NP> by <NP NO="4" CorefID="4">Charles E. Raymond, District Attorney</NP>. <NP NO="5" CorefID="5">Ierulli</NP> will replace <NP NO="6" CorefID="6">Desmond D. Connall</NP>  who has been called to <NP NO="7" CorefID="7">active military service</NP> but is expected back on <NP NO="8" CorefID="8">the job</NP> by <NP NO="9" CorefID="9">March 31</NP>.\\\hline

\end{tabulary}

\textbf{Table 2}: Results of applying five publicly available automated NER and/or co-reference resolution tools on a portion of Brown Corpus text.
\pagebreak

It must be stressed that, exploitation of produced output for the text segmentation task requires identification of named entity mentions as well as co-reference mentions. This observation also holds for all the aforementioned tools. Table 2 lists the output obtained from the six tools examined in a small portion of a Brown corpus text appearing in Choi’s dataset. Examination of the obtained output leads to the following observations:

\begin{enumerate}
\item In order to avoid manual annotation, an already trained model for Named Entity Recognition and co-reference resolution should be used. Best results are achieved when the model is trained in a related topic(s). For the problem examined, models trained in the MUC corpus are preferred. MUC is either a Message Understanding Conference or a Message Understanding Competition. At the sixth conference (MUC-6) the task of recognition of named entities and co-reference was added. The majority of publicly available tools, used as training data documents provided in that competition (\url{http://www.cs.nyu.edu/cs/faculty/grishman/muc6.html}), including Illinois and Reconcile tools, as stated in (\cite{{Stoyanov:2010}, {Ratinov:2009}}). By the term “\textit{best results}”, we mean high accuracy in recognizing named entity instances as well as mentions resulting from co-reference resolution measured by metrics such as Precision and Recall.
\item The usefullness and quality of the obtained result/output is highly related to: a) the types of named entities that the model is trained to (thus, the NER tool is able to attribute); b) the types of co-reference that is able to recognize (anaphora resolution, pronoun resolution etc); c) potential manual correction and completion of the outcome for NER and co-reference resolution for the needs of the problem in question; d) attribution of a unique named entity identifier to each distinct instance (including all mentions of it). This means that, the quality of produced output depends on the quality of the output of a number of individual parts. This is in alignment with Atdağ and Labatut (\cite{Atdag:2013}) statements as well as Appelt (\cite{Appelt:1999}) who states that: '\textit{Two factors must be taken under consideration: a) the Information extraction task is based on a number of preprocessing steps such as tokenization, sentence splitting, shallow parsing etc, and is divided into a number of sub-tasks such as Named Entity Recognition, Template Element Task, Template Relation Task, Coreference resolution etc.; b) the Information extraction task follows either the knowledge based approach … or supervised learning based approach where a large annotated corpus is available to provide examples on which learning algorithms can operate}.' This implies that, portability and suitability of an already existing information extraction system is highly dependent on the suitability of any of its constituent parts as well as whether each of them is domain or vocabulary dependent. Moreover, Marrero et al. (\cite{Marrero:2009}) claim that: '\textit{The number of categories (i.e., named entity types) that each tool can recognize is an important factor for the evaluation of a tool… In other words, an important factor in the evaluation of the different systems is not only the number of different entity types recognized but also their “quality}'. Metrics presenting the average performance in the identification of entity types is not always representative of its success.”
\item The suitability of the produced output on whether it can be given as input to a text segmentation algorithm presents strong variation. Initial examination of publicly available tools proved that an important number of them produce an output in xml format, without providing a parser to transform xml to text format. Additionally, other tools choose to represent notions, corresponding for example to noun phrases or chunking, in a unique manner. In each of those cases, a dedicated parser must be constructed in order to transform the produced output in a form that can be processed by a text segmentation algorithm.
\item An important drawback of those tools is that, some provide a dedicated component such as co-reference (for example Reconcile or Guitar) while others, provide more complete solutions (such as Stanford NLP Group) but produce output in xml format.
\item In all cases, post-processing is necessary in order: a) to correct mistakes or enhance the output with named entity instances or co-reference resolution outcome; b) to add named entity instance identifier (the same to all related named entity instances within the same text); c) to transform the output into a form that can be given as input to a text segmentation algorithm (the construction of an dedicated parser may here be required). 
\end{enumerate}

The problem thus is in finding the correct tool or combination of tools (trained with the most thematically related corpus) that produce a reliable output, and perform the appropriate post-processing. To this direction we performed a second type of annotation (apart from the manual one described in the sequel) by using publicly available automated annotation tools. More specifically, this annotation involved use of two combinations: a) use of Illinois NER tool and Illinois Co-reference tool; b) use of Illinois NER tool and Reconcile Co-referencer as well as use of Illinois NER tool only. It must be stressed that, during construction of the aforementioned tools, MUC dataset was used for training (\cite{{Stoyanov:2010}, {Ratinov:2009}}).

It is worth mentioning that, the construction of the parser first of all requires the identification of all rules / patterns under which each automated tool attribute the information (i.e., named entity instance or co-reference mention). More specifically, Illinois NER proves to fail to attribute correct named entity types to documents belonging to category J of the Brown Corpus. A possible explanation to this can be that, category J contains scientific documents consisting of technical terms. Those terms are usually tagged as Group name instances by SemCor, while they are usually characterized as PERSON or ORGANIZATION by Illinois NER. The fact that the current version used here is able to recognize 19 different named entity types may have an impact in the statistical distribution of named entity instances within a portion of a text i.e., a segment. This reveals the second problem that occurred, which is the type of the information that is captured by each tool and the way that is represented in combination with others.

Reconcile Co-referencer fails to follow a unique rule in the attribution of a noun phrase which has a severe impact in finding the correct mention of a named entity instance. Reconcile Co-referencer fails to detect mentions of the same named entity instance. Additionally, both Reconcile and Illinois Co-referencers fail to detect pronominal anaphora, in other words, the same identifier is attributed to all words inside a text (such who, whome, your, etc.) but this identifier is not associated with the corresponding named entity instance.

To the best of the author's knowledge all published work for text segmentation takes as input plain text and not text in other data form such as xml, csv etc. NER and co- refererence resolution tools are applied in plain text before applying any text segmentation algorithm, in order to identify named entities and all mentions of them. Another step before applying any text segmentation algorithm is the application of a parser constructed by the author in order to attribute the same named entity identifier to a named entity instance and its related instances i.e., mentions.

The two combinations of NER and co-reference systems produced a complicated output which cannot be given as is as input to a text segmentation algorithm. Thus, a dedicated parser was manually constructed for each of them. Additionally, the aforementioned constructed parsers attribute a unique named entity identifier to each entity mention.

Since, as it was proved earlier, the selection of an efficient combination of Named Entity Recognition and co-reference resolution tools along with the construction of a dedicated parser to post-process the produced output, and the application of them to the problem examined presents a considerable complexity, we performed manual NER and co-reference resolution on each of the ten segments of the 700 samples. In order to cover  the majority of named entities and mentions in each segment, we selected four types of named entities: person name, location, date, and group name. The most general type is that of group name, which is used for the annotation of words and terms not falling into the other categories. It was also used for the annotation of scientific terms frequently appearing in segments.

Attention must be paid to the fact that in Semcor (\url{http://multisemcor.itc.it/semcor.php}), a different annotation for the majority of documents belonging to categories A and J was performed. The English SemCor corpus is a sense-tagged corpus of English created at Princeton University by the WordNet Project research team (\cite{Landes:1998}) and is one of the first sense-tagged corpora produced for any language. The corpus consists of a subset of the Brown Corpus (700,000 words, with more than 200,000 sense-annotated), and it has been part-of- speech-tagged and sense-tagged. For each sentence, open class words (or multi-word expressions) and named entities are tagged. Not all expressions are tagged. Most specifically, '\textit{The Semcor corpus is composed of 352 texts. In 186 texts, all open class words (nouns, adjectives, and adverbs) are annotated with PoS, lemma and sense according to Princenton Wordnet 1.6, while in the remaining 166 texts only verbs are annotated with lemma and sense}'. This type of annotation differs from the one performed here. Even though in Semcor nouns are classified into three categories (person name, group name, and location), substitution of every reference of the same instance with the same named entity identifier as a result of the identification of identical named entities and the application of co-reference resolution, is not performed. Additionally, Semcor does not provide annotations for all documents belonging to category J, nor for all named entity instances (for example, scientific terms such as \textit{urethane foam}).

Taking under consideration the deficiencies of Semcor, we performed in each segment of Choi's corpus manual annotation of proper names belonging to one of the four categories. The annotation took under consideration the assignment of lemmas to categories of person name, group name, and location appearing in Semcor. However, it is our belief that substitution of words with named entity instances does not have a negative effect in the performance of a text segmentation algorithm. More specifically, we expect that co-reference resolution will reinforce mentions of named entity instances whose frequency of appearance in greater to one, since they are not elimitated as the result of stop list removal and stemming. This is the reason why, during manual named entity annotation we paid special attention to those pronouns that correspond to named entity instances. More specifically, we additionally: (a) substituted every reference of the same instance with the same named entity id. For example in the sentences '\textit{James P. Mitchell and Sen. Walter H. Jones R-Bergen, last night disagreed on the value of using as a campaign issue a remark by Richard J. Hughes,... . Mitchell was for using it, Jones against, and Sen. Wayne Dumont Jr ....}', we first identified four instances of person names. We further used the same entity id for \textit{James P. Mitchell} and \textit{Mitchell}, and the same entity id for \textit{Sen. Walter H.Jones R-Bergen} and \textit{Jones}; (b) we substituted every reference of the same instance resulted from co-reference resolution with the same named entity id (for example in the sentences '\textit{Mr. Hawksley, the state's general treasurer,...... He is not interested in being named a full-time director}.', we substituted \textit{He} with the named entity id given to \textit{Mr. Hawksley}). 

It must be stressed that, the named entity type of Organization was not used in Semcor annotation since it is rather restricting for the topics mentioned in Brown Corpus. Group name was chosen instead as the 'default' named entity type, in order to cover not only scientific terms but also other document specific types. For specific areas, it covers the notion of organization while for others, such as those covering scientific areas, it is used to cover scientific terms. Group name was used in SemCor and was preserved for compatibility reasons. Group name is also used as a named entity type in Libre Multilingual Analyzer (\url{http:://aymara.github.io/lima}). It must be stressed that, in scientific documents, the author noticed that instances corresponding to organization named entity type do not appear, thus no confusion comes into view in the use of group name entity type. It is an assumption that was made. Examination of more named entity types is considered as future work.

In align with Secmor, group names involved expressions such as '\textit{House Committee on Revenue and Taxation}' or '\textit{City Executive Committee}'. The annotation of location instances included possible derivations of them such as '\textit{Russian}'. The annotation of date instances included both simple date form (consisting only of the year or month) and more complex forms (containing both month, date, and year). It must be stressed that co- reference resolution was performed only on portions of text that refer to entity instances and not on the text as a whole. For the purposes of the current paper manual annotation was performed as well as a post-processing step for error correction, identification of all named entity mentions, and attribution of the same entity identifier to each distinct named entity instance.

The annotation process revealed that segments belonging to category A contain on average more named entity instances compared to those belonging to category J. The difference in the results is highly related to the topic discussed in segments of each category. More specifically, for each of the 124 documents of the Brown Corpus we selected the largest part used in Choi's benchmark (from the original corpus i.e., not the one produced after annotation) as segment i.e., portions of eleven sentences. We then counted the minimum, maximum, and average number of named entity instances appearing in them (as the result of the manual annotation process). The results are listed in Table 3. Since the annotation was performed by a single annotator, the kappa statistic that measures the inter-annotator agreement cannot be calculated.

\begin{center}%
\begin{tabular}
[c]{|r|r|r|r|}\hline
\textbf{Category/ NE instances per segment } & \textbf{Min} & \textbf{Max} & \textbf{Average}\\\hline
Segments of Category A & 2 & 53 & 28.3\\\hline
Segments of Category J & 2 & 57 & 18.4\\\hline
\end{tabular}

\textbf{Table 3}:Statistics regarding  the  number of  named  entity  instances appearing  in  segments of Category A and J.
\end{center}

The concatenated texts produced in Choi's dataset differ from real texts in the sense that each concatenated text is a sample of ten texts. One would expect that, each segment to have a limited number of named entities, which would consequently influence the application of co-reference resolution. On the other hand, a real text contains significantly more entities as well as re-appearances of them within it. However, observation of our concatenated texts (as it can be seen from the example listed in Table 4) proved that it is not the case. The reason for this is the appearance of an important number of named entity instances as well as pronouns within a segment. An example of the annotation process is listed in Table 4 in which A21 prefix corresponds to the fact that this paragraph was taken from document 21 belonging to Category A of the Brown Corpus. From the  example above it is obvious that identical named entity instances as well as mentions corresponding to pronouns are substituted with the same identifier.

\begin{tabulary}{\textwidth}{|L|L|}\hline
\textbf{Non annotated paragraph} & \textbf{Annotated paragraph} \\\hline
St. Johns , Mich. , April 19 . A jury of seven men and five women found 21-year- old Richard Pohl guilty of manslaughter yesterday in the bludgeon slaying of Mrs. Anna Hengesbach. Pohl received the verdict without visible emotion. He returned to his cell in the county jail, where he has been held since his arrest last July, without a word to his court-appointed attorney, Jack Walker, or his guard. Stepson vindicated The verdict brought vindication to the dead woman 's stepson, Vincent Hengesbach, 54, who was tried for the same crime in December, 1958, and released when the jury failed to reach a verdict. Mrs. Hengesbach was killed on Aug. 31, 1958. Hengesbach has been living under a cloud ever since. When the verdict came in against his young neighbor, Hengesbach said: `` I am very pleased to have the doub of suspicion removed. Still , I don't wish to appear happy at somebody else 's misfortune''. Lives on welfare Hengesbach, who has been living on welfare recently, said he hopes to rebuild the farm which was settled by his grandfather in Westphalia, 27 miles southwest of here. Hengesbach has been living in Grand Ledge since his house and barn were burned down after his release in 1958. 
& 
A21location1, A21location2, A21date1. A jury of seven men and five women found 21-year-old A21person1 guilty of manslaughter yesterday in the bludgeon slaying of A21person2. A21person1 received the verdict without visible emotion. A21person1 returned to A21person1 cell in the county jail, where A21person1 has been held since A21person1 arrest last A21date2, without a word to A21person1 court- appointed attorney, A21person3, or A21person1 guard. A21person4 vindicated The verdict brought vindication to the A21person2's stepson, A21person5, 54, A21person5 was tried for the same crime in A21date3, and released when the jury failed to reach a verdict. A21person5 was killed on A21date4. A21person5 has been living under a cloud ever since. When the verdict came in against A21person5 young neighbor, A21person5 said : `` A21person5 am very pleased to have the doubt of suspicion removed. Still, A21person5 don't wish to appear happy at somebody else's misfortune ''. Lives on welfare Hengesbach A21person5, A21person5 has been living on welfare recently, said A21person5 hopes to rebuild the farm which was settled by A21person5 grandfather in A21location3, 27 miles southwest of here. A21person5 has been living in A21location5 since A21person5 house and barn were burned down after A21person5 release in 1958 A21date5.
 \\\hline
\end{tabulary}
\linebreak

\textbf{Table 4}:Portion of a segment belonging to Choi' s benchmark, before and after performing manual NER and co-reference resolution.\\
\linebreak

Table 5 provides the output of each type of annotation performed (manual annotation, using Illinois NER alone, using Illinois NER along with each of the two Co- referencers). In this example, A13 prefix corresponds to the fact that this paragraph was taken from document 13 belonging to Category A of the Brown Corpus.

\begin{tabulary}{1\textwidth}{|L|L|}\hline
\small{Non annotated paragraph} & \small {Rookie Ron Nischwitz continued his pinpoint pitching Monday night as the Bears made it two straight over Indianapolis, 5-3.The husky 6-3, 205-pound lefthander, was in command all the way before an on-the-scene audience of only 949 and countless of television viewers in the Denver area.It was Nischwitz' third straight victory of the new season and ran the Grizzlies' winning streak to four straight.They now lead Louisville by a full game on top of the American Association pack}.\\\hline
\small{Automatically Annotated paragraph with Illinois NER} & \small{
Rookie [PERSON Ron Nischwitz] continued his pinpoint pitching [TIME Monday night] as the Bears made it [CARDINAL two] straight over [GPE Indianapolis],5-3.The husky 6-3, 205-pound lefthander, was in command all the way before an on-the-scene audience of only h[CARDINAL 949] and countless of television viewers in the [GPE Denver] area.It was [PERSON Nischwitz]'[ORDINAL third] straight victory of the new season and ran the Grizzlies' winning streak to [CARDINAL four] straight.They now lead [GPE Louisville] by a full game on top of [ORG the American Association] pack}.\\\hline
\small{Automatically Annotated paragraph with Illinois NER and Reconcile Co-referencer} & \tiny {<NP NO="119" CorefID="120">Rookie [PERSON Ron Nischwitz</NP>] continued <NP NO="121" CorefID="123"><NP NO="120" CorefID="120">his</NP> pinpoint pitching [TIME Monday night</NP>] as <NP NO="122" CorefID="122">the Bears</NP> made <NP NO="123" CorefID="123">it</NP><NP NO="124" CorefID="124">[CARDINAL two</NP>] straight over <NP NO="125" CorefID="135">[GPE Indianapolis</NP>],<NP NO="126" CorefID="126">5-3</NP>.<NP NO="127" CorefID="127">The husky 6-3</NP>,<NP NO="128" CorefID="128">205-pound lefthander</NP>, was in <NP NO="129" CorefID="129">command</NP> <NP NO="130" CorefID="130">all the way</NP> before <NP NO="131" CorefID="131">an on-the-scene audience of <NP NO="132" CorefID="132">only [CARDINAL 949] and <NP NO="133" CorefID="133">countless</NP> of <NP NO="134" CorefID="134">television viewers</NP> in <NP NO="135" CorefID="135">the [GPE <NP NO="136" CorefID="136">Denver</NP>] area</NP></NP></NP>.<NP NO="137" CorefID="143">It</NP> was [PERSON Nischwitz]'<NP NO="138" CorefID="138">[ORDINAL third] straight victory of <NP NO="139" CorefID="139">the new season</NP></NP> and ran <NP NO="141 CorefID="141"><NP NO="140" CorefID="140">the Grizzlies</NP>'winning streak</NP> to <NP NO="142" CorefID="142">[CARDINAL four</NP>] straight.<NP NO="143" CorefID="143">They</NP> now lead <NP NO="144" CorefID="144">[GPE</NP> <NP NO="145" CorefID="145">Louisville</NP>] by <NP NO="146" CorefID="146">a full game on <NP NO="147" CorefID="147">top</NP> of <NP NO="148" CorefID="148">[ORG</NP></NP> <NP NO="149" CorefID="149">the American Association</NP>] pack} .\\\hline
\small{Automatically Annotated paragraph with Illinois NER \& Co-referencer} & \small {\*Rookie\*\_28 [\*\*PERSON\*\_75 Ron Nischwitz\*\_21] continued \*his\*\_21 pinpoint pitching [TIME Monday night] as \*the Bears\*\_75 made it [\*\*CARDINAL\*\_47 two*\_21] straight over [\*GPE Indianapolis\*\_42], 5\-3.The husky 6\-3 , \*205\-pound lefthander\*\_30, was in command all the way before \*an on-the-scene audience of only [\*CARDINAL\*\_47 949\*\_9] and countless of \*television viewers\*\_7 in \*the [\*GPE Denver\*\_42] area\*\_67.It was [\*PERSON Nischwitz\*\_64] \'[ORDINAL third] straight victory of the new season and ran \*the Grizzlies\*\_50 \'winning streak to [\*\*CARDINAL\*\_47 four*\_47] straight.\*They\*\_0 now lead [\*GPE Louisville\*\_42] by a full game on top of [\*ORG the American Association\*\_34] pack}. \\\hline
\small{Manually Annotated paragraph} &  \small{A13person1 continued A13person1 pinpoint pitching A13date1 night as the A13group1 made it two straight over A13location1 , 5-3 . The husky 6-3, 205-pound lefthander, was in command all the way before an on-the-scene audience of only 949 and countless of television viewers in the A13location2 area. It was A13person1' third straight victory of the new season and ran the A13group2' winning streak to four straight.A13group2 now lead A13location3 by a full game on top of the A13group3 pack.A13person1 fanned six and walked only A13person2 in the third inning. A13person1 has given only the one pass in his 27 innings, an unusual characteristic for a southpaw.The A13group1 took the lead in the first inning,as A13group1 did in A13date2's opener, and never lagged. A13person3 cracked the first of his two doubles against A13person4 to open the Bear A13group1's attack.After A13person5 gruonded out,A13person6 walked and A13person7 singled home A13person3. A13person8 then moved A13person6 across with a line drive to left. A13person9 drew a base on balls to fill the bases but A13person10's smash was knocked down by A13person4 for the the American Association pack}.\\\hline
\end{tabulary}
\linebreak
\textbf{Table 5}:Portion of a segment belonging to Choi' s benchmark, before and after performing manual NER and co-reference resolution (portion of document br-a13 of the Brown Corpus), annotation provided by Semcor as well as using automated annotation tools.

\section{Greek Corpus}

For our experiments, we used the corpus created in (\cite{Fragkou:2007}). There, the authors used a text collection compiled from Stamatatos Corpus (\cite{Stamatatos:2001}), comprising of text downloaded from the website of the newspaper 'To Vima' (\url{http://tovima.dolnet.gr}). Stamatatos, Fakorakis and Kokkinakis (\cite{Stamatatos:2001}) constructed a corpus collecting texts like essays on Biology, Linguistics, Archeology, Culture, History, Technology, Society, International Affairs, and Philosophy from ten different authors. Thirty texts were selected from each author. Table 6 lists the authors contributing to Stamatatos Corpus as well as the thematic area(s) covered by each of them.

\begin{center}%
\begin{tabular}
[c]{|r|r|}\hline
\textbf{Author } & \textbf{Thematic Area }\\\hline
Alachiotis & Biology\\\hline
Babiniotis & Linguistics\\\hline
Dertilis & History,Society\\\hline
Kiosse & Archeology\\\hline
Liakos & History,Society\\\hline
Maronitis & Culture,Society\\\hline
Ploritis & Culture,History\\\hline
Tassios & Technology,Society\\\hline
Tsukalas & International Affairs\\\hline
Vokos & Philosophy\\\hline
\end{tabular}

\textbf{Table 6}: List of Authors and Thematic Areas dealt by each of those.
\end{center}

In the work presented in {\cite{Fragkou:2007}), each of the 300 texts of the collection of articles compiled from the newspaper 'To Vima' was pre-processed using the POS tagger developed in {\cite{Orphanos:1999}. The tagger is based on a lexicon and is capable of assigning full morphosyntactic attributes to 876,000 Greek word forms. In experiments presented in  {\cite{Fragkou:2007}), every noun, verb, adjective, or adverb in the text was substituted by its lemma, as determined by the tagger. For those words that their lemma was not determined by the tagger, no substitution was made. The authors in  {\cite{Fragkou:2007})created two groups of experiments (which are described in details in Section 7.2); their difference lies in the length of the created segments and the number of authors used for the creation of the texts to segment. Each text was a concatenation of ten text segments. Each author was characterized by the vocabulary he uses in his texts. Hence, the goal in  {\cite{Fragkou:2007}) was to segment the text into the parts written by each author. The previously described corpus was also used for our experiments due to its uniqueness to the problem examined.

\subsection{Named Entity Annotation}

There exist a limited number of readily-available automated annotation tools in the literature. For our experiments, we used the corpus created in  {\cite{Fragkou:2007}). More specifically, we applied the automated annotation tool described in \cite{Lucarelli:2007}. This annotation tool was chosen because it is publicly available, it was trained on similar documents taken from the newspaper 'Ta Nea', and produces an output that can be easily be given as input to a text segmentation algorithm. Newspaper 'Ta Nea' contains articles having similar content with that of the newspaper 'To Vima'. The annotation tool was thus applied in our corpus without requiring training. We have chosen four types of named entities i.e., person name, group name, location, and date. The annotation tool produced annotations for some, but not all instances of person names, group names, and dates. In order to annotate all named entities appearing in each text, a second pass was performed. During this pass, manual completion of named entity annotation of proper names belonging to one of the four categories was performed in each segment. No correction was performed, since the annotation tool was proven to perform correct annotation to those named entity instances that could identify and appropriately classify. In alignment with the English corpus, during manual completion of named entity annotation, we additionally: (a) annotated all instances of locations and (b) substituted every reference of the same instance with the same named entity identifier i.e., performed co-reference resolution to identify all mentions that represent the same entity and grouping them to the entity they refer to, by paying special attention to the appearance of Greek pronouns. The latter step was necessary because the annotation tool used cannot perform co-reference resolution. It must be stressed that, no parser was needed to be constructed since the produced output was in a form that could be easily be given as input to a segmentation algorithm. 


The annotation process led to the conclusion that, texts having a social subject usually contain a small number of named entity instances, contrary to texts about politics, science, archeology, history, and philosophy. For example, texts belonging to the author Kiosse contain on average big number of named entities, because they describe historical events issuing person names, dates, and locations. Table 7 provides an overview of the average number of named entity instances appearing in the annotated documents for each author in the corpus. Once again, since manual completion of annotation was performed by a single annotator, the kappa statistic which measures the inter-annotator agreement cannot be calculated.

\begin{center}%
\begin{tabular}
[c]{|r|r|}\hline
\textbf{Author } & \textbf{Average number of NE's}\\\hline
Alachiotis & 44.00\\\hline
Babiniotis & 70.23\\\hline
Dertilis & 33.33\\\hline
Kiosse & 121.90\\\hline
Liakos & 77.70\\\hline
Maronitis & 40.40\\\hline
Ploritis & 94.20\\\hline
Tassios & 40.00\\\hline
Tsukalas & 37.12\\\hline
Vokos &52.16\\\hline
\end{tabular}

\textbf{Table 7}: Statistics regarding the average number of named entity instances appearing in the annotated documents of the Stamatatos Corpus per author.
\end{center}

\section{Text Segmentation Algorithms}

The annotated corpora that resulted from the previously described named entity annotation process during which, every word or phrase was firstly categorized to one of the predefined named entity types and secondly was substituted by a unique named entity identifier, were evaluated using five text segmentation algorithms for English and four for Greek texts respectively. The first is Choi's C99b, which divides the input text into minimal units on sentence boundaries and computes a similarity matrix for sentences based on the cosine similarity (\cite{Choi:2001}). Each sentence is represented as a T- dimensional vector, where T denotes the number of topics selected for the topic model, where each element of this vector contains the number of times a topic occurs in a sentence. Next, a rank matrix R is computed by calculating for each element of the similarity matrix, the number of neighbors of it that have lower similarity scores than itself. As a final step, a top-down hierarchical clustering algorithm is performed to divide the document into segments recursively, by splitting the ranking matrix according to a threshold-based criterion i.e., the gradient decent along the matrix diagonal.

The second algorithm is the one proposed by Utiyama and Isahara (\cite{Utiyama:2001}). This algorithm finds the optimal segmentation of a given text by defining a statistical model, which calculates the probability of words belonging to a segment. Utiyama and Isahara's algorithm models each segment using the conventional multinomial model, assuming that segment specific parameters are estimated using the usual maximum likelihood estimates with Laplace smoothing. The second term intervening in the probability of a segmentation, is the penalty factor. Utiyama and Isahara's algorithm  (\cite{Utiyama:2001}) principle is to search globally for the best path in a graph representing all possible segmentations, where edges are valued according to the lexical cohesion measured in a probabilistic way. Both algorithms have the advantage that they do not require training and their implementation is publicly available.

The third algorithm used is introduced by Kehagias et al. (\cite{Kehagias:2004}). In contrary to the previous ones, it requires training. More specifically, this algorithm uses dynamic programming to find both the number as well as the location of segment boundaries. The algorithm decides the locations of boundaries by calculating the globally optimal splitting (i.e., global minimum of a segmentation cost) on the basis of a similarity matrix, a preferred fragment length, and a cost function defined. High segmentation accuracy is achieved in cases where produced segments length present small derivation from the actual segment length. High segment length means greater number of sentences appearing in it, which augments the probability of obtaining an important number of words and possibly named entity instances appearing in it.

Two additional algorithms, whose code is publicly available were used for comparison purposes. The first is the Affinity Propagation algorithm introduced by Kazantseva and Szpakowicz ({\cite{Kazantseva:2011}), while the second is MinCutSeg introduced by Malioutov and Barzilay (\cite{Malioutov:2006}). Affinity propagation algorithm takes as input measures of similarity between pairs of data points and simultaneously considers all data points as potential exemplars. Real- valued messages are exchanged between data points until a high-quality set of exemplars and corresponding clusters gradually emerges. For the segmentation task, Affinity Propagation creates segment centers and segment assignment for each sentence.

On the other hand, MinCutSeg treats segmentation as a graph-partitioning task that optimizes the normalized cut criterion. More specifically, Malioutov and Barzilay's criterion measures both the similarity within each partition and the dissimilarity across different partitions. Thus, MinCutSeg not only considers localized comparisons but also takes into account long-range changes in lexical distribution.

\section{Evaluation Metrics}

The evaluation of the algorithms both in the original and annotated corpora was performed using four widely known metrics: Precision, Recall, Beeferman's Pk (\cite{Beeferman:1999}), and WindowDiff  ({\cite{Pevzner:2002}). Precision is defined as '\textit{the number of the estimated segment boundaries which are actual segment boundaries' divided by 'the number of the estimated segment boundaries}'. Recall is defined as '\textit{the number of the estimated segment boundaries which are actual segment boundaries' divided by 'the number of the true segment boundaries}'. Beeferman's Pk metric measures the proportion of '\textit{sentences which are wrongly predicted to belong to different segments (while they actually belong in the same segment)}' or '\textit{sentences which are wrongly predicted to belong to the same segment (while they actually belong in different segments)}'. Beeferman's Pk is a window-based metric which attempts to solve the harsh near-miss penalization of Precision, Recall, and F-measure. In Beeferman's Pk, a window of size k - where k is defined as half of the mean reference segment size - is slid across the text to compute penalties. A penalty of 1 is assigned for each window whose boundaries are detected to be in different segments of the reference and hypothesis segmentations, and this count is normalized by the number of windows. It is worth noticing that, Beeferman's Pk metric measures segmentation inaccuracy. Thus, small values of the metric correspond to high segmentation accuracy.

A variation of Beeferman's \textit{Pk} metric named WindowDiff index was proposed by Pevzner and Hearst  ({\cite{Pevzner:2002})which remendies several of Beeferman's \textit{Pk} problems. Pevzner and Hearst highlighted a number of issues with Beeferman's \textit{Pk}, specifically that: i) False negatives (FNs) are penalized more than false positives (FPs); ii) Beeferman's \textit{Pk} does not penalize FPs that fall within k units of a reference boundary; iii) Beeferman's \textit{Pk}'s sensitivity to variations in segment size can cause it to linearly decrease the penalty for FPs if the size of any segments fall below k; and iv) Near-miss errors are too harshly penalized. To attempt to mitigate the shortcomings of Beeferman's\textit{ Pk}, Pevzner and Hearst ({\cite{Pevzner:2002}) proposed a modified metric which changed how penalties were counted, named WindowDiff. A window of size k is still slid across the text, but now penalties are attributed to windows, where the number of boundaries in each segmentation differs with the same normalization. WindowDiff is able to reduce, but not eliminate, sensitivity to segment size, gives more equal weights to both FPs and FNs (FNs are in effect penalized less), and is able to catch mistakes in both small and large segments. Lamprier et al. (\cite{Lamprier:2007}) demonstrated that WindowDiff penalizes errors less at the beginning and end of a segmentation.

Recent work in evaluation metrics includes the work of Scaiano and Inkpen (\cite{Scaiano:2012}). The authors introduce WinPR, which resolves some of the limitations of WindowDiff. WinPR distinguishes between false positive and false negative errors as the result of a confusion matrix; is insensitive to window size, which allows us to customize near miss sensitivity; and is based on counting errors not windows, but still provides partial reward for near misses. WinPR counts boundaries, not windows, which has analytical benefits.

Finally, Kazantseva and Szpakowicz (\cite{Kazantseva:2012}) also proposed a simple modification to WindowDiff which allows for taking into account, more than one reference segmentations, and thus rewards or penalizes the output of automatic segmenters by  considering the severity of their mistakes. However, the proposed metric is a window- based metric, so its value depends on the choice of the window size. The metric also hides whether false positives or false negatives are the main source of error and is based on inter-annotation agreement.

Since Precision, Recall, Beeferman's \textit{Pk} metric and WindowDiff are the most widely used evaluation metrics, they are also used in the experiments listed below.

\section{Experiments}

Experiments were performed to examine the impact of substituting words or phrases with named entity instances in the performance of a text segmentation algorithm. Towards this direction, one group of experiments was performed for English and two groups for Greek. Subsection 7.1 presents the experiments performed in the English corpus, while Subsection 7.2 presents those performed in the Greek corpus. All experiments used the segmentation algorithms presented in Section 5 and the evaluation metrics described in Section 6.

\subsection{Experiments in English Corpus}

Complementary to annotation, stop-word removal and stemming was performed (i.e., suffix removal) based on Porter's algorithm (\cite{Porter:1980}) before applying the segmentation algorithms in the original and the manually annotated corpus. Stop-word removal and stemming was also performed in the output produced after applying information extraction tools used in our experiments. Comparison of the contribution of named entity annotation is provided in Table 8. Table 8 contains the results obtained after applying the segmentation algorithms of Choi’s C99b, Utiyama \& Isahara, Kehagias et al., Affinity Propagation, and MinCutSeg in: Case 1) the original corpus; Case 2) the manully annotated Choi's Corpus; Case 3) the corpus produced after applying Illinois NER and Reconcile Co-referencer automated annotation tools; Case 4) the corpus produced after applying Illinois NER and Illinois Co-referencer automated annotation tools; Case 5) the corpus produced by applying Illinois NER only. In Case 5 post processing took place after applying Illinois NER tool in order to firstly remove all unnecessary tags (such as brackets etc.) and secondly attribute an identifier to each unique named entity instance. Regarding Affinity Propagation, available source code does not provide segmentation accuracy measured by Beeferman's \textit{Pk} metric. Table 9 which results from Table 8, provides the difference in performance accuracy (measured using Beeferman's \textit{Pk} and WindowDiff metrics) between any of the four different types of annotation (i.e., manual, using Illinois NER only, using Illinois NER along with Illinois co-referencer or Reconcile co-referencer) and manual annotation, for all algorithms and all datasets. In Table 8 bold notation denotes the best performance obtained by Window Diff metric in all cases when examing a specific segmentation algorithm. 
Table 10 lists the average number of NE instances after performing manual and automatic annotation. In Table 10 we can see that for any annotation type, the lowest average number of named entity instances occurs in Set (3-5) where segment length varies from 3 to 5 sentences. On the other hand the highest average number of named entity instances occurs in Set (9-11) where segment length varies from 9 to 11 sentences. This implies that high segment length favors the appearance of more named entity instances.

It is worth pointing that, reported results in the literature regarding text segmentation, provide very small variations to obtained performance measured using Beeferman's \textit{Pk} and WindowDiff metrics. Thus, statistical significance cannot be calculated. Moreover, to the best of the author’s knowledge, statistical significance in the bibliography related to text segmentation is not usually calculated. Moreover, since this is a preliminary study, from the conducted experiments a firm conclusion regarding statistical significance cannot be drawed.

\begin{landscape}
\begin{tabulary}{1.30\textwidth}
{|l|l|l|l|l|l|l|l|l|l|l|l|l|l|l|l|l|l|l|l|l|l|}\\\hline

\tiny{Algorithm} & \tiny{Dataset} & \tiny{Precis} & \tiny{Precis} & \tiny{Precis} & \tiny{Precis} & \tiny{Precis} & \tiny{Rec} & \tiny{Rec} & \tiny{Rec} & \tiny{Rec} & \tiny{Rec} & \tiny{PK} & \tiny{PK} & \tiny{PK} & \tiny{PK} & \tiny{PK} & \tiny{WinDiff} & \tiny{WinDiff} & \tiny{WinDiff} & \tiny{WinDiff} & \tiny{WinDiff} \\\hline
\tiny{} & \tiny{} & \tiny{Case 1} & \tiny{Case 2} & \tiny{Case 3} & \tiny{Case 4} & \tiny{Case 5} & \tiny{Case 1} & \tiny{Case 2} & \tiny{Case 3} & \tiny{Case 4} & \tiny{Case 5} & \tiny{Case 1} & \tiny{Case 2} & \tiny{Case 3} & \tiny{Case 4} & \tiny{Case 5} & \tiny{Case 1} & \tiny{Case 2} & \tiny{Case 3} & \tiny{Case 4} & \tiny{Case 5} \\\hline
\tiny{Choi's C99b} & \tiny{Set1(3-11)} & \tiny{78} & \tiny{81.84} & \tiny{59.36} & \tiny{58.43} & \tiny{61.91} & \tiny{78} & \tiny{81.84} & \tiny{59.49} & \tiny{57.80} & \tiny{63.36} & \tiny{12.1} & \tiny{10.87} & \tiny{16.91} & \tiny{20.64} & \tiny{15.41} & \tiny{12.90} & \tiny{\textbf{11.61}} & \tiny{19.58} & \tiny{21.34} & \tiny{16.07}\\\hline
\tiny{Choi's C99b} & \tiny{Set2(3-5)} & \tiny{85.6} & \tiny{89.75} & \tiny{69.7} & \tiny{62.4} & \tiny{59.78} & \tiny{85.6} & \tiny{89.75} & \tiny{68.82} & \tiny{63.38} & \tiny{63.38} & \tiny{10.4} & \tiny{8.61} & \tiny{14.96} & \tiny{17.93} & \tiny{14.71} & \tiny{10.71} & \tiny{\textbf{8.86}} & \tiny{15.82} & \tiny{18.65} & \tiny{15.17}\\\hline
\tiny{Choi's C99b} & \tiny{Set3(6-8)} & \tiny{80.7} & \tiny{85.62} & \tiny{66.9} & \tiny{60.1} & \tiny{49.7} & \tiny{80.7} & \tiny{85.62} & \tiny{66.58} & \tiny{60.96} & \tiny{55.99} & \tiny{7} & \tiny{8.42} & \tiny{14.16} & \tiny{17.91} & \tiny{13.55} & \tiny{9.54} & \tiny{\textbf{8.59}} & \tiny{14.83} & \tiny{18.39} & \tiny{14.03}\\\hline
\tiny{Choi's C99b} & \tiny{Set4(9-11)} & \tiny{86.5} & \tiny{86.25} & \tiny{67.6} & \tiny{60.7} & \tiny{54.8} & \tiny{86.5} & \tiny{86.25} & \tiny{66.80} & \tiny{61.95} & \tiny{61.85} & \tiny{8.5} & \tiny{8.11} & \tiny{13.94} & \tiny{17.26} & \tiny{11.11} & \tiny{8.62} & \tiny{\textbf{8.32}} & \tiny{14.64} & \tiny{16.81} & \tiny{11.45}\\\hline
\tiny{Choi's C99b} & \tiny{All Files} & \tiny{80.7} & \tiny{84.14} & \tiny{63.09} & \tiny{59.56} & \tiny{58.84} & \tiny{80.7} & \tiny{84.14} & \tiny{62.88} & \tiny{59.64} & \tiny{62.09} & \tiny{11} & \tiny{9.80} & \tiny{15.82} & \tiny{19.38} & \tiny{14.43} & \tiny{11.49} & \tiny{\textbf{10.32}} & \tiny{17.66} & \tiny{19.89} & \tiny{14.99}\\\hline
\tiny{Utiyama} & \tiny{Set1(3-11)} & \tiny{67.40} & \tiny{79.47} & \tiny{67.87} & \tiny{44.96} & \tiny{73.36} & \tiny{70.63} & \tiny{74.55} & \tiny{66.44} & \tiny{55.20} & \tiny{61.81} & \tiny{13.85} & \tiny{11.49} & \tiny{16.31} & \tiny{26.61} & \tiny{13.25} & \tiny{\textbf{12.27}} & \tiny{13.71} & \tiny{20.39} & \tiny{35.65} & \tiny{15.47}\\\hline
\tiny{Utiyama} & \tiny{Set2(3-5)} & \tiny{77.81} & \tiny{82.26} & \tiny{67.64} & \tiny{40.42} & \tiny{48.44} & \tiny{74.19} & \tiny{79.63} & \tiny{67.53} & \tiny{57.44} & \tiny{45.65} & \tiny{9.99} & \tiny{8.21} & \tiny{16.38} & \tiny{30.40} & \tiny{25.39} & \tiny{9.79} & \tiny{\textbf{8.57}} & \tiny{18.97} & \tiny{43.04} & \tiny{26.39}\\\hline
\tiny{Utiyama} & \tiny{Set3(6-8)} & \tiny{77.87} & \tiny{90.66} & \tiny{72.26} & \tiny{27.58} & \tiny{77.81} & \tiny{86.70} & \tiny{90.66} & \tiny{76.22} & \tiny{63.92} & \tiny{70.71} & \tiny{3.51} & \tiny{2.45} & \tiny{10.00} & \tiny{35.96} & \tiny{6.66} & \tiny{3.39} & \tiny{\textbf{2.34}} & \tiny{13.18} & \tiny{57.88} & \tiny{6.98}\\\hline
\tiny{Utiyama} & \tiny{Set4(9-11)} & \tiny{79.31} & \tiny{87.55} & \tiny{67.83} & \tiny{23.43} & \tiny{70.52} & \tiny{87.75} & \tiny{87.11} & \tiny{73.74} & \tiny{53.34} & \tiny{67.69} & \tiny{3.24} & \tiny{3.30} & \tiny{10.83} & \tiny{35.93} & \tiny{9.66} & \tiny{\textbf{3.16}} & \tiny{3.24} & \tiny{14.84} & \tiny{55.32} & \tiny{12.15}\\\hline
\tiny{Utiyama} & \tiny{All Files} & \tiny{72.08} & \tiny{82.62} & \tiny{68.46} & \tiny{38.75} & \tiny{70.02} & \tiny{75.88} & \tiny{79.37} & \tiny{69.03} & \tiny{56.50} & \tiny{61.61} & \tiny{10.3} & \tiny{8.56} & \tiny{14.63} & \tiny{29.82} & \tiny{13.53} & \tiny{\textbf{9.34}} & \tiny{9.85} & \tiny{18.36} & \tiny{42.69} & \tiny{15.34}\\\hline
\tiny{Kehagias} & \tiny{Set1(3-11)} & \tiny{72.61} & \tiny{73.10} & \tiny{65.11} & \tiny{53.44} & \tiny{62.22} & \tiny{70.88} & \tiny{70.66} & \tiny{62.65} & \tiny{53.77} & \tiny{50.85} & \tiny{11.73} & \tiny{11.81} & \tiny{15.54} & \tiny{19.16} & \tiny{17.89} & \tiny{\textbf{12.80}} & \tiny{\textbf{12.80}} & \tiny{17.04} & \tiny{21.12} & \tiny{19.20}\\\hline
\tiny{Kehagias} & \tiny{Set2(3-5)} & \tiny{83.88} & \tiny{83.88} & \tiny{71.83} & \tiny{58.97} & \tiny{54.02} & \tiny{81.77} & \tiny{81.77} & \tiny{70.56} & \tiny{57.77} & \tiny{46.65} & \tiny{7.08} & \tiny{7.08} & \tiny{12.34} & \tiny{19.27} & \tiny{22.41} & \tiny{\textbf{7.03}} & \tiny{\textbf{7.03}} & \tiny{12.70} & \tiny{20.45} & \tiny{23.70}\\\hline
\tiny{Kehagias} & \tiny{Set3(6-8)} & \tiny{87.77} & \tiny{89.25} & \tiny{79.14} & \tiny{62.24} & \tiny{68.98} & \tiny{87.77} & \tiny{89.11} & \tiny{79.56} & \tiny{64.78} & \tiny{59.86} & \tiny{2.57} & \tiny{2.67} & \tiny{5.96} & \tiny{12.96} & \tiny{13.80} & \tiny{\textbf{2.49}} & \tiny{2.77} & \tiny{6.37} & \tiny{13.82} & \tiny{14.97}\\\hline
\tiny{Kehagias} & \tiny{Set4(9-11)} & \tiny{86.66} & \tiny{87.77} & \tiny{79.55} & \tiny{65.48} & \tiny{62.83} & \tiny{86.66} & \tiny{87.77} & \tiny{77.89} & \tiny{66.42} & \tiny{61.14} & \tiny{1.96} & \tiny{1.77} & \tiny{5.93} & \tiny{10.89} & \tiny{13.90} & \tiny{\textbf{1.88}} & \tiny{\textbf{1.88}} & \tiny{6.376} & \tiny{11.94} & \tiny{15.41}\\\hline
\tiny{Kehagias} & \tiny{All Files} & \tiny{78.40} & \tiny{79.05} & \tiny{70.14} & \tiny{57.21} & \tiny{62.10} & \tiny{77.11} & \tiny{77.33} & \tiny{68.37} & \tiny{57.72} & \tiny{53.00} & \tiny{8.36} & \tiny{8.40} & \tiny{12.34} & \tiny{17.11} & \tiny{17.38} & \tiny{\textbf{8.94}} & \tiny{8.98} & \tiny{13.37} & \tiny{18.67} & \tiny{18.70}\\\hline
\tiny{Affinity Prop} & \tiny{Set1(3-11)} & \tiny{10.86} & \tiny{11.01} & \tiny{10.72} & \tiny{11.08} & \tiny{12.21} & \tiny{7.78} & \tiny{8.08} & \tiny{7.86} & \tiny{13.67} & \tiny{22.61} & \tiny{-} & \tiny{-} & \tiny{-} & \tiny{-} & \tiny{-} & \tiny{31.57} & \tiny{31.33} & \tiny{\textbf{19.58}} & \tiny{26.19} & \tiny{50.41}\\\hline
\tiny{Affinity Prop} & \tiny{Set2(3-5)} & \tiny{18.14} & \tiny{17.89} & \tiny{17.39} & \tiny{15.96} & \tiny{20.59} & \tiny{12.5} & \tiny{12.16} & \tiny{12.24} & \tiny{11.10} & \tiny{12.63} & \tiny{-} & \tiny{-} & \tiny{-} & \tiny{-} & \tiny{-} & \tiny{17.12} & \tiny{\textbf{16.89}} & \tiny{40.92} & \tiny{47.82} & \tiny{17.84}\\\hline
\tiny{Affinity Prop} & \tiny{Set3(6-8)} & \tiny{9.75} & \tiny{11.05} & \tiny{11.29} & \tiny{10.33} & \tiny{11.84} & \tiny{7.5} & \tiny{8.5} & \tiny{8.65} & \tiny{13} & \tiny{12.36} & \tiny{-} & \tiny{-} & \tiny{-} & \tiny{-} & \tiny{-} & \tiny{35.00} & \tiny{34.84} & \tiny{34.38} & \tiny{32.43} & \tiny{\textbf{28.78}}\\\hline
\tiny{Affinity Prop} & \tiny{Set4(9-11)} & \tiny{15.95} & \tiny{11.99} & \tiny{12.24} & \tiny{10.92} & \tiny{8.48} & \tiny{14.33} & \tiny{11.67} & \tiny{11.22} & \tiny{19.83} & \tiny{12.81} & \tiny{-} & \tiny{-} & \tiny{-} & \tiny{-} & \tiny{-} & \tiny{\textbf{34.55}} & \tiny{34.91} & \tiny{34.75} & \tiny{50.06} & \tiny{45.96}\\\hline
\tiny{Affinity Prop} & \tiny{All Files} & \tiny{12.47} & \tiny{12.14} & \tiny{11.97} & \tiny{11.65} & \tiny{12.82} & \tiny{9.35} & \tiny{9.23} & \tiny{9.08} & \tiny{14.09} & \tiny{18.32} & \tiny{-} & \tiny{-} & \tiny{-} & \tiny{-} & \tiny{-} & \tiny{30.42} & \tiny{\textbf{30.28}} & \tiny{36.06} & \tiny{42.85} & \tiny{42.03}\\\hline
\tiny{MinCutSeg} & \tiny{Set1(3-11)} & \tiny{24.76} & \tiny{25.64} & \tiny{22.04} & \tiny{17.70} & \tiny{18.48} & \tiny{23.22} & \tiny{24.27} & \tiny{21.25} & \tiny{16.47} & \tiny{16.52} & \tiny{26.57} & \tiny{25.40} & \tiny{40.75} & \tiny{42.14} & \tiny{32.90} & \tiny{29.64} & \tiny{\textbf{28.46}} & \tiny{44.98} & \tiny{46.70} & \tiny{37.08}\\\hline
\tiny{MinCutSeg} & \tiny{Set2(3-5)} & \tiny{20.62} & \tiny{20.90} & \tiny{21.95} & \tiny{23.10} & \tiny{20.72} & \tiny{18.33} & \tiny{18.5} & \tiny{19.66} & \tiny{20.5} & \tiny{17} & \tiny{40.58} & \tiny{40.14} & \tiny{34.74} & \tiny{41.14} & \tiny{42.83} & \tiny{44.40} & \tiny{43.77} & \tiny{\textbf{39.10}} & \tiny{45.72} & \tiny{48.43}\\\hline
\tiny{MinCutSeg} & \tiny{Set3(6-8)} & \tiny{24.80} & \tiny{31.24} & \tiny{21.96} & \tiny{21.06} & \tiny{19.27} & \tiny{23.33} & \tiny{30} & \tiny{21.21} & \tiny{19.5} & \tiny{17.7} & \tiny{28.45} & \tiny{27.06} & \tiny{30.26} & \tiny{34.51} & \tiny{32.84} & \tiny{30.01} & \tiny{\textbf{28.48}} & \tiny{32.89} & \tiny{37.18} & \tiny{36.46}\\\hline
\tiny{MinCutSeg} & \tiny{Set4(9-11)} & \tiny{28.68} & \tiny{31.34} & \tiny{21.61} & \tiny{20.17} & \tiny{21.60} & \tiny{27.67} & \tiny{30.67} & \tiny{21.48} & \tiny{19.5} & \tiny{20.5} & \tiny{22.15} & \tiny{20.28} & \tiny{23.29} & \tiny{28.36} & \tiny{26.55} & \tiny{22.97} & \tiny{\textbf{20.82}} & \tiny{24.79} & \tiny{29.92} & \tiny{28.96}\\\hline
\tiny{MinCutSeg} & \tiny{All Files} & \tiny{24.74} & \tiny{26.57} & \tiny{21.96} & \tiny{19.31} & \tiny{19.36} & \tiny{23.17} & \tiny{25.18} & \tiny{21.05} & \tiny{17.91} & \tiny{17.32} & \tiny{28.21} & \tiny{27.01} & \tiny{33.32} & \tiny{38.51} & \tiny{33.40} & \tiny{30.85} & \tiny{\textbf{29.56}} & \tiny{37.01} & \tiny{42.38} & \tiny{37.45}\\\hline
\end{tabulary}
\textbf{Table 8}: Performance of the five segmentation algorithms applied on the original Choi's Corpus for English.Case 1 denotes obtained results in the non-annotated corpus, Case 2 in the manually annotated corpus, Case 3 in the coprus produced using Illinois NER and Reconcile Co-referencer, Case 4 in the coprus produced using Illinois NER and Co-referencer, while Case 5 in the coprus produced using Illinois NER only without performing co-reference resolution.
\end{landscape}

\begin{landscape}

\begin{tabulary}{1.30\textwidth}
{|l|l|l|l|l|l|l|l|l|l|} \\\hline
\tiny{\textbf{Algorithm }} & \tiny{\textbf{Dataset }} & \tiny{\textbf{PK man ann}} & \tiny{\textbf{PK Illinois NER}} & \tiny{\textbf{PK Illinois NER}} & \tiny{\textbf{PK Illinois NER}} & \tiny{\textbf{WinDiff man ann}} & \tiny{\textbf{WinDiff Illinois NER}} & \tiny{\textbf{WinDiff Illinois NER}} & \tiny{\textbf{WinDiff Illinois NER}}\\\hline
\tiny{\textbf{}} & \tiny{\textbf{}} & \tiny{\textbf{- PK non an}} & \tiny{\textbf{\& Reconcile}} & \tiny{\textbf{\& coref}} & \tiny{\textbf{Only}} & \tiny{\textbf{- WinDiff non ann}} & \tiny{\textbf{\& Reconcile}} & \tiny{\textbf{\& co-ref}} & \tiny{\textbf{Only}}\\\hline
\tiny{\textbf{}} & \tiny{\textbf{}} & \tiny{\textbf{}} & \tiny{\textbf{- PK non ann}} & \tiny{\textbf{- PK
non ann}} & \tiny{\textbf{- PK non ann}} & \tiny{\textbf{}} & \tiny{\textbf{- WinDiff non ann}} & \tiny{\textbf{- WinDiff non ann}} & \tiny{\textbf{- WinDiff non ann}}\\\hline
\tiny{Choi's C99b} & \tiny{Set1(3-11)} & \tiny{-1.23} & \tiny{4.81} & \tiny{8.54} & \tiny{3.31} & \tiny{-1.29} & \tiny{6.68} & \tiny{8.44} & \tiny{3.17}\\\hline
\tiny{Choi's C99b} & \tiny{Set2(3-5)} & \tiny{-1.79} & \tiny{4.56} & \tiny{7.53} & \tiny{4.31} & \tiny{-1.85} & \tiny{5.11} & \tiny{7.94} & \tiny{4.46}\\\hline
\tiny{Choi's C99b} & \tiny{Set3(6-8)} & \tiny{1.42} & \tiny{7.16} & \tiny{10.91} & \tiny{6.55} & \tiny{-0.95} & \tiny{5.29} & \tiny{8.85} & \tiny{4.49}\\\hline
\tiny{Choi's C99b} & \tiny{Set4(9-11)} & \tiny{-0.39} & \tiny{5.44} & \tiny{8.76} & \tiny{2.61} & \tiny{-0.3} & \tiny{6.02} & \tiny{8.19} & \tiny{2.83}\\\hline
\tiny{Choi's C99b} & \tiny{All Files} & \tiny{-1.2} & \tiny{4.82} & \tiny{8.38} & \tiny{3.43} & \tiny{-1.17} & \tiny{6.17} & \tiny{8.4} & \tiny{3.5}\\\hline
\tiny{Utiyama} & \tiny{Set1(3-11)} & \tiny{-2.36} & \tiny{2.46} & \tiny{12.76} & \tiny{-0.6} & \tiny{1.44} & \tiny{8.12} & \tiny{23.38} & \tiny{3.2}\\\hline
\tiny{Utiyama} & \tiny{Set2(3-5)} & \tiny{-1.78} & \tiny{6.39} & \tiny{20.41} & \tiny{15.4} & \tiny{-1.22} & \tiny{9.18} & \tiny{33.25} & \tiny{16.6}\\\hline
\tiny{Utiyama } & \tiny{Set3(6-8)} & \tiny{-1.06} & \tiny{6.49} & \tiny{32.45} & \tiny{3.15} & \tiny{-1.05} & \tiny{9.79} & \tiny{54.49} & \tiny{3.59}\\\hline
\tiny{Utiyama} & \tiny{Set4(9-11)} & \tiny{0.06} & \tiny{7.59} & \tiny{32.69} & \tiny{6.42} & \tiny{0.08} & \tiny{11.68} & \tiny{52.16} & \tiny{8.99}\\\hline
\tiny{Utiyama} & \tiny{All Files} & \tiny{-1.74} & \tiny{4.33} & \tiny{19.52} & \tiny{3.23} & \tiny{0.51} & \tiny{9.02} & \tiny{33.35} & \tiny{6}\\\hline
\tiny{Kehagias} & \tiny{Set1(3-11)} & \tiny{0.08} & \tiny{3.81} & \tiny{7.43} & \tiny{6.16} & \tiny{0} & \tiny{4.24} & \tiny{8.32} & \tiny{6.4}\\\hline
\tiny{Kehagias} & \tiny{Set2(3-5)} & \tiny{0} & \tiny{5.26} & \tiny{12.19} & \tiny{15.33} & \tiny{0} & \tiny{5.67} & \tiny{13.42} & \tiny{16.67}\\\hline
\tiny{Kehagias} & \tiny{Set3(6-8)} & \tiny{0.1} & \tiny{3.39} & \tiny{10.39} & \tiny{11.23} & \tiny{0.28} & \tiny{3.88} & \tiny{11.33} & \tiny{12.48}\\\hline
\tiny{Kehagias} & \tiny{Set4(9-11)} & \tiny{-0.19} & \tiny{3.97} & \tiny{8.93} & \tiny{11.94} & \tiny{0} & \tiny{6374.12} & \tiny{10.06} & \tiny{13.53}\\\hline
\tiny{Kehagias.} & \tiny{All Files} & \tiny{0.04} & \tiny{3.98} & \tiny{8.75} & \tiny{9.02} & \tiny{0.04} & \tiny{4.43} & \tiny{9.73} & \tiny{9.76}\\\hline
\tiny{Affinity Prop} & \tiny{Set1(3-11)} & \tiny{\-} & \tiny{\-} & \tiny{\-} & \tiny{\-} & \tiny{-0.24} & \tiny{-11.99} & \tiny{-5.38} & \tiny{18.84}\\\hline
\tiny{Affinity Prop} & \tiny{Set2(3-5)} & \tiny{\-} & \tiny{\-} & \tiny{\-} & \tiny{\-} & \tiny{-0.23} & \tiny{23.8} & \tiny{30.7} & \tiny{0.72}\\\hline
\tiny{Affinity Prop} & \tiny{Set3(6-8)} & \tiny{\-} & \tiny{\-} & \tiny{\-} & \tiny{\-} & \tiny{-0.16} & \tiny{-0.62} & \tiny{-2.57} & \tiny{-6.22}\\\hline
\tiny{Affinity Prop} & \tiny{Set4(9-11)} & \tiny{\-} & \tiny{\-} & \tiny{\-} & \tiny{\-} & \tiny{0.36} & \tiny{0.2} & \tiny{15.51} & \tiny{11.41}\\\hline
\tiny{Affinity Prop} & \tiny{All Files} & \tiny{\-} & \tiny{\-} & \tiny{\-} & \tiny{\-} & \tiny{-0.14} & \tiny{5.64} & \tiny{12.43} & \tiny{11.61}\\\hline
\tiny{MinCutSeg} & \tiny{Set1(3-11)} & \tiny{-1.17} & \tiny{14.18} & \tiny{15.57} & \tiny{6.33} & \tiny{-1.18} & \tiny{15.34} & \tiny{17.06} & \tiny{7.44}\\\hline
\tiny{MinCutSeg} & \tiny{Set2(3-5)} & \tiny{-0.44} & \tiny{-5.84} & \tiny{0.56} & \tiny{2.25} & \tiny{-0.63} & \tiny{-5.3} & \tiny{1.32} & \tiny{4.03}\\\hline
\tiny{MinCutSeg} & \tiny{Set3(6-8)} & \tiny{-1.39} & \tiny{1.81} & \tiny{6.06} & \tiny{4.39} & \tiny{-1.53} & \tiny{2.88} & \tiny{7.17} & \tiny{6.45}\\\hline
\tiny{MinCutSeg} & \tiny{Set4(9-11)} & \tiny{-1.87} & \tiny{1.14} & \tiny{6.21} & \tiny{4.4} & \tiny{-2.15} & \tiny{1.82} & \tiny{6.95} & \tiny{5.99}\\\hline
\tiny{MinCutSeg} & \tiny{AllFiles} & \tiny{-1.2} & \tiny{5.11} & \tiny{10.3} & \tiny{5.19} & \tiny{-1.29} & \tiny{6.16} & \tiny{11.53} & \tiny{6.6}\\\hline
\end{tabulary}

\textbf{Table 9}: Differences in performance obtained by the five segmentation algorithms applied on the original Choi's Corpus for English (measured using Beeferman's Pk and WindowDiff metrics) between any of the four different types of annotation (i.e., manual, using Illinois NER only, using Illinois NER along with Illinois co-referencer or Reconcile co-referencer) and non- annotation.
\end{landscape}

\begin{tabulary}{1.25\textwidth}
{|l|l|l|l||l|}\\\hline
\small\textbf{Average number} & \small\textbf{Manual } & \small\textbf{Illinois NER only} & \small\textbf{Illinois NER } & \small\textbf{Illinois NER\& Recon-} \\\hline
\small\textbf{of NE instances} & \small\textbf{Annotation} & \small\textbf{} & \small\textbf{ \& Co-referencer} & \small\textbf{ cile Co-referencer} \\\hline
Set(3-11) & 130.89 & 94.48 & 195.19 & 107.72\\\hline
Set(3-5) & 71.31 & 51.35 & 111.81 & 64.04\\\hline
Set(6-8) & 134.4 & 91.95 & 200.11 & 116.02\\\hline
Set(9-11) & 180.16 & 128.87 & 271.76 & 145\\\hline
\end{tabulary}

\textbf{Table 10}: Average Number of NE instances after performing manual and automatic annotation.

The general conclusion of the conducted experiments is that, all segmentation algorithms work better with manual annotation, (compared to original raw texts). Moreover, segmentation results obtained with automatic annotations strongly depend on the tools used and are sometimes worse, even worse than using just the raw texts. Additionally, combination of Illinois NER and Reconcile co-referencer proves to be more effective than combination of Illinois NER and co-referencer. This does not provide a clear view regarding the contribution of co-reference due to the nature of each automatic tool.

More specifically, regarding manual annotation, results shown in Tables 8 and 9 lead to the following observation: a significant improvement was obtained in all measures and for all datasets regarding Choi's C99b algorithm. An exception to this statement is the slight decrease of Precision and Recall in Set4 (9-11) from 86.5 per cent (in the original corpus) to 86.25 per cent (in the annotated corpus). Difference in performance in all measures and for all datasets also holds for the results obtained after applying the algorithm of Utiyama \& Isahara, especially in Set3 (6-8) (with a small exception in Set4 (9-11)). Amelioration is achieved in all measures and for all datasets after applying MinCutSeg algorithm especially in Set4 (9-11). It demostrates that the algorithm performs better when the segment's length is high, since the lowest improvement appears in Set2 (3-5). Better performance is also achieved in Set4 (9-11) after applying the Affinity Propagation algorithm measured by WindowDiff in the manually annotated corpus. For the rest of datasets difference in performance and metrics varies. The Kehagias et al. algorithm fails to obtain better performance in the first three datasets. Additionally, Precision and Recall resulted from Kehagias et al. algorithm do show any improvement for sets Set3 (6-8) and Set4 (9-11). On the contrary, in Set4 (9-11), the difference in the -already high- performance is marginal. This is an indication that the algorithm performs better when the segment's length is high and the derivation of the expected segment length is small. The greater difference is observed in Set3 (6-8) and Set4 (9-11) for the majority of algorithms. This can be justified by the fact that, in those datasets the number of named entity instances and those resulting after co-reference resolution in manual annotation is higher than the equivalent in the previous ones, as shown in Table 10. Special attention must be given to co-reference resolution, which showed to have played an important role in the variation of the number of named entity instances per segment in manual annotation.

It is worth mentioning that, both combinations of automated tools exhibit the same “behavior” regarding the performance obtained by various algorithms. By the term “\textit{behavior}”, we mean here the same performance achieved in terms of: (a) the algorithm used for segmenting texts; and (b) every dataset in which each of the aforementioned algorithms was applied to. For both combinations, best performance is achieved by Kehagias et al., Choi’s and Utihama and Isahara’s algorithms respectively, while the worst performace is achieved firstly by Affinity Propagation and secondly by MinCutSeg algorithm. More specifically, combination of Illinois NER and Reconcile Co-rerefencer exhibits the same behavior as the manual annotated corpus (when compared with performance obtained in non-annotated texts) in every subset in Choi's, Utiyama and Isahara's as well as Kehagias et al. algorihtms. Once again, Choi's, Utiyama and Isahara's as well as Kehagias et al. algorithms perform better than Affinity Propagation and MinCutSeg algorithm. Additionally, Choi's, Utiyama and Isahara's as well as Kehagias et al. algorithms, best performance, measured using Beeferman's Pk and WindowDiff metrics is observed in Set3 (6-8) and Set4 (9-11).

On the other hand, combination of Illinois NER and Illinois Co-rerefencer exhibits the same behavior as the manual annotated corpus only in Kehagias et al. segmentation algorithm. Overall, both combinations obtain worse performance that manual annotation in all algorithms and for all datasets.

Illinois NER and Reconcile Co-referencer combination shows to perform better than the one of Illinois automated annotation tools, in all algorithms, for all datasets and for all metrics apart from Precision and Recall values in Set (3-11) for both Affinity Propagation and MinCutSeg algorihtms. The difference in performance between those combinations is worse in the performance obtained by Utiyama and Isahara's algorithm for all metrics and all datasets.

Regarding the performance obtained by using automatic annotation tools compared with raw i.e., non- annotated texts we can see that, obtained results are overall worse with small exceptions.
Regarding the role of co-reference resolution tools in performance achieved instead of using only Illinois NER, observation of information included in Tables 8 and 8 show that, contribution of co-reference seems to be influenced by the tool used. It appears that, Reconcile co-referencer is more effective than the one of Illinois. Its contribution is apparent in Kehagias et al. and MinCutSeg algorithms, in almost all metrics and all sets. However, combination for Illinois NER and co-referencer performs overall better than Illinois NER only in Kehagias et al. algorithm. Even though, as it can be easily be seen from Table 10, the number of named entity instances produced after applying Illinois reference is greater than those produced after applying Reconcile co- referencer, the latter proves to be more efficient. This can be attributed to the “quality”of named entity instances produced as well as how each segmentation algorithm profits them.
Obtained results from the use of automated annotation tools can be attributed to the
following factors: a) the outcome of co-reference resolution from both tools; b) the suitability and validity of the constructed parsers, which deserves further investigation; c) the combination of tools was not an ideal one; d) the appropriateness of the corpus in which each tool was trained to; e) the number of named entity types and how statistical distribution of words is affected.

The degree to which automatic NER and co-reference systems may produce efficient results is strongly related to the following factors: a) the nature of the dataset for segmentation used; b) the NER system, which ideally should be trained on similar corpus, which means that the domain of the dataset used for segmentation and the equivalent the NER system was trained to, are highly related and ideally must be as close as possible, since this affects the accuracy of produced predictions; c) the number of named entity types used by the NER tool as well as whether they are generic or focus only on a specific domain, since this highly affects the number of named entity instances they can capture and subsequently the statistical distribution of terms (i.e., words and named entity instances) that results from their application. Depending on the nature of the NER tool (too generic or domain specific) there is a risk of failing to capture named entity instances or to attribute an important number of them to the “default” named entity type appearing in the majority of NER tools; d) the types of co-reference that a co-reference system captures, since there exist a number of different types of co-reference and every co- reference system is trained in order to be able to capture some or (rarely) all types; e) the validity of the constructed parser to tranform a "tagged" output resulted from the application of a NER and a co-reference system to plain text, which is necessary in order to be given as input to a text segmentation algorithm. The correctness of the parser is highly affected from the implementation of the NER tool used i.e., the ability to detect tagging rules used by every system. Table 5 helps us to understand the diversity of different outputs.

Our claims are in align with Atdağ and Labatut (\cite{Atdag:2013}), Appelt (\cite{Appelt:1999}), Siefkes (\cite{Siefkes:2007}) and Marrero et al. (\cite{Marrero:2009}) statements, which are presented in Section 3.1. An important observation is that named entity type instance acts indirectly as a discriminative factor in the segmentation process. More specifically for manual annotation, dominant types are person name, since person names appear more frequently, and group name, since this named entity type is used for the annotation of words and terms not falling into other categories. This leads to segments containing named entity instances whose majority belongs to the aforementioned types. Complementary, co- reference resolution reinforces mentions of named entity instances, since they are not eliminated as the result of stop list removal and stemming. Thus resulting segments contain more named entity instances than those produced after performing word pre- processing. The use of limited number of named entity types in the manual annotation confirm to reinforce semantic coherence compared to the use of the 19 named entity types of Illinois NER.

It must be stressed that, named entity type is strongly related to the document's topic. Thus, appropriate selection of named entity types, according to the document's topic may lead to correct attribution of instances to named entity types, affecting named entity distribution and consequently intra and inter segment similarity. For example, a document belonging to category A which contains documents about Political, Sports, Society, Sport News, Financial, and Cultural is expected to have more named entity instances of person and group name than date or location. The frequency of appearance of those named entities is reinforced by co-reference, thus affects word distribution given as input to a text segmentation algorithm. Word distribution is also affected by the fact that named entity instances are frequently expressions containing more than one word, a portion of which is usually eliminated after performing stop word removal and stemming. All the aforementioned observations proved to affect segmentation since the frequency of appearance of named entity instances is increased thus, reinforcing intra-segment similarity. This observation holds both for English and Greek language. The whole rational is in contrast with information extraction, where the learning process takes into account only the type of named entities occurring in a passage of text i.e., when a named entity recognizer is trained in order to be able to identify from an already annotated corpus new named entity instances in a testing dataset.

In the work presented in Alajmi et al. (\cite{Alajmi:2012}) the authors claim that '\textit{the elimination of stop words reduces the corpus size typically by 20 to 30\%}'. This assumption is reinforced by Savoy (\cite{Savoy:1999}) who claims that, '\textit{The elimination of stop words also reduces the corpus size typically by 30\% and thus leads to higher efficiency}'.

\subsection{Experiments in Greek Corpus}

The expectation that substitution of words with named entity instances does not have a negative impact in the performance of text segmentation algorithms was also examined for Greek texts, by performing two groups of experiments. For both groups, Orphanos and Christodoulakis (\cite{Orphanos:1999}) POS tagger was applied, leading to selection of lemmas that are either nouns or verbs or adjectives or adverbs. For those words that their lemma was not determined by the tagger, no substitution was made.

\subsubsection{First Group of Experiments}

As it was previously mentioned, we use the Greek corpus examined in  (\cite{Fragkou:2007}), consisting of portions of texts taken from the Greek newspaper 'To Vima'. The collection of texts used for the first group of experiments, consists of six datasets: Set0,..., Set5. Each of those datasets differs in the number of authors used for the generation of the texts to segment and consequently in the number of texts used from the entire collection (\cite{Stamatatos:2001}), as depicted in Table 11.

\begin{center}%
\begin{tabular}
[c]{|r|r|r|}\hline
\textbf{Dataset } & \textbf{Authors } & \textbf{No. of docs per set}\\\hline
Set0 & Kiosse,Alachiotis & 60\\\hline
Set1 & Kiosse,Maronitis & 60\\\hline
Set2 & Kiosse,Alachiotis,Maronitis & 90\\\hline
Set3 & Kiosse,Alachiotis,Maronitis,Ploritis & 120\\\hline
Set4 & Kiosse,Alachiotis,Maronitis,Ploritis,Vokos & 150\\\hline
Set5 & All Authors & 300\\\hline
\end{tabular}

\textbf{Table 11}: List of the sets compiled in the first group of experiments using Greek texts and the author’s texts used for each of them.
\end{center}

For each of the above datasets, four subsets were constructed. Those subsets differ in 
the number of sentences appearing in each segment. Let Lmin and Lmax be the smallest 
and largest number of sentences that a segment may contain. Four different ( Lmin , Lmax ) pairs were used: (3,11); (3,5); (6,8); and (9,11). Hence, Set0 contains four subsets: Set01, Set02, Set03 and similarly for Set1, Set2, ..., Set5. The datasets Set*1 are the ones with ( Lmin , Lmax ) = (3,11), the datasets Set*2 are the ones with ( Lmin , Lmax ) =(3,5), and so on. Let also { X1 ,..., X n } be the authors contributing to the generation of the dataset. Texts  belonging in each dataset are generated by the following procedure:
Each text is the concatenation of ten segments. For each segment we do the following:
\begin{itemize}
\item  An author from { X1 ,..., X n } is randomly selected. Let I be the selected author.
\item  A text among the thirty available that belong to the I author is randomly selected. Let k be the selected text of author I.
\item  A number l in (Lmin, Lmax) ∈ is randomly selected.
\item  Finally, l consecutive sentences from text k (starting at the first sentence of the text) are extracted. Those sentences constitute the generated segment.
\end{itemize}

In order to validate our expectation, we applied Choi’s C99b, Utiyama \& Isahara, Kehagias et al., and Affinity Propagation segmentation algorithms as before in all datasets. MinCutSeg algorithm failed to provide (reliable) segmentation results. More specifically, MinCutSeg configuration file requires the existence of a stop list and a stemmer. Use of available to the author Greek stop list and stemmer proved to be incompatible to the algorithm's specifications. Moreover, MinCutSeg fails to recognize reference boundary annotations. It must be stressed that, MinCutSeg is trained in an English corpus, which justifies potential failure of recognizing segment boundaries appearing in a different language due to encoding. On the other hand, Affinity Propagation fails to produce results for both the annotated and non-annotated corpus for Set*2 (3-5). It seems that the algorithm fails to identify exemplars i.e., cluster centers for every examined document. This can be attributed to the small number of sentences appearing in Set*2 (3-5), the choice of parameter values (especially the 'preference' parameter) as well as problems encountered with similarity measurement yielding to oscillations i.e., non convergence. Problems regarding similarity measurement as well as parameter selection are addressed in subsequent implementations of the algorithm.  

Table 12 lists the values of Precision, Recall, Beeferman's PK and WindowDiff  metrics reported in the literature, after applying Choi’s C99b, Utiyama \& Isahara's, Kehagias et al., and Affinity Propagation algorithms: (a) in the non- annotated corpus; (b) in the manual annotated corpus; (c) in the corpus resulting after applying Lucarelli's et al. annotation tool only, averaged over all datasets which have segments of the same length as well as those obtained after applying the same algorithms on the same datasets where annotation was previously performed. 

We reach the following conclusions based on the obtained results. Affinity Propagation algorithm's performance is -marginally- better in the annotated corpus in all datasets. Regarding the algorithm of Utiyama \& Isahara, a significant improvement was obtained in all measures and for all datasets of this group of experiments i.e., for manual annotation. This can be justified by the fact that Utiyama \& Isahara's algorithm performs global optimization of local information. In contrast, Choi's C99b and Kehagias et al. algorithms perform local optimization of global information and global optimization of global information respectively. The same observation holds for the results obtained after applying Choi's C99b and Kehagias et al. algorithms, where improvement can be observed in all evaluation metrics and for all datasets and for manual annotation. This  improvement appears to be greater in datasets Set*1 (3-11) and Set*2 (3-5) in all algorithms. This is an indication that the annotation succeeded in identifying critical information which, in other ways, was lost. For datasets Set*3 (6-8) and Set*4 (9-11), the segmentation accuracy remains high. The reason for this is that in those datasets, the segment length is high leading to a high number of named entity instances.

Co-reference resolution significantly improves segmentation accuracy than using NE annotation only. This statement holds for all algorithms and for all datasets (except for precision and recall performance obtained by Affinity Propagation algorithm in Set*3 (6-8) and Set*4 (9-11)). Use of Lucarelli's annotation tool only compared to non- annotated texts fails to improve segmentation accuracy.

\pagebreak
\begin{landscape}
\begin{tabulary}{1.35\textwidth}
{|l|l|l|l|l|l|l|l|l|l|l|l|l|l|}\\\hline
\small{\textbf{Algorithm}} & \small{\textbf{Dataset}} &\small{\textbf{Precis}} & \small{\textbf{Precis}} & \small{\textbf{Precis}} &\small{\textbf{Recall}} & \small{\textbf{Recall}} & \small{\textbf{Recall}} &\small{\textbf{Pk}} & \small{\textbf{Pk}} & \small{\textbf{Pk}}&\small{\textbf{WindowDiff}} & \small{\textbf{WindowDiff}} & \small{\textbf{WindowDiff}}\\\hline
\textbf{} &\textbf{} &\small \textbf{no} &\small \textbf{NES} & \small\textbf{with} &\small\textbf{no} & \small\textbf{NES} & \small\textbf{with} &\small\textbf{no} & \small\textbf{NES} & \small\textbf{with}&\small\textbf{no} & \small\textbf{NES} & \small\textbf{with}\\\hline
\textbf{} &\textbf{} & \small\textbf{annotat} & \small\textbf{annotat} & \small\textbf{annotat} &\small\textbf{annotat} & \small\textbf{annotat} & \small\textbf{annotat} &\small\textbf{annotat} & \small\textbf{annotat} & \small\textbf{annotat}&\small\textbf{annotat} & \small\textbf{annotat} & \small\textbf{annotat}\\\hline
\small{Choi's C99b} & \small{Set*1(3-11)} & \small{59.7} & \small{48.73} & \small{63.26} & \small{59.67} & \small{44.51} & \small{63.26} & \small{17.96} & \small{19.54} & \small{15.96} & \small{19.37} & \small{21.34} & \small{17.40}\\\hline
\small{Choi's C99b} & \small{Set*1(3-5)} & \small{67.86} & \small{52.9} & \small{70.46} & \small{67.86} & \small{46.94} & \small{70.46} & \small{16.70} & \small{20.82} & \small{14.53} & \small{17.93} & \small{22.93} & \small{15.91}\\\hline
\small{Choi's C99b} & \small{Set*1(6-8)} & \small{64.9} & \small{54.57} & \small{71.26} & \small{64.9} & \small{54.56} & \small{71.26} & \small{15.13} & \small{14.06} & \small{11.92} & \small{15.89} & \small{15.17} & \small{12.45}\\\hline
\small{Choi's C99b} & \small{Set*1(9-11)} & \small{64.23} & \small{53.1} & \small{68.46} & \small{64.23} & \small{48.8} & \small{68.46} & \small{13.60} & \small{16.30} & \small{11.43} & \small{14.07} & \small{16.71} &\small{11.89}\\\hline
\small{Choi's C99b} & \small{All Files)} & \small{64.17} & \small{52.32} & \small{68.36} & \small{64.17} & \small{48.70} & \small{68.36} & \small{15.85} & \small{17.68} & \small{13.46} & \small{16.82} & \small{19.04} & \small{14.41}\\\hline
\small{Utiyama} & \small{Set*1(3-11)} & \small{64.18} & \small{51.69} & \small{70.74} & \small{61.24} & \small{44.99} & \small{66.96} & \small{17.47} & \small{18.38} & \small{13.72} & \small{18.48} & \small{19.50} & \small{14.70}\\\hline
\small{Utiyama} & \small{Set*1(3-5)} & \small{70.04} & \small{54.37} & \small{76.65} & \small{54.74} & \small{35.68} & \small{61.55} & \small{20.99} & \small{26.24} & \small{16.83} & \small{21.31} & \small{26.71} & \small{17.30}\\\hline
\small{Utiyama} & \small{Set*1(6-8)} & \small{75.45} & \small{55.45} & \small{80.31} & \small{73.07} & \small{52.74} & \small{78.18} & \small{10.96} & \small{14.00} & \small{8.43} & \small{11.00} & \small{14.07} & \small{8.44}\\\hline
\small{Utiyama} & \small{Set*1(9-11)} & \small{73.17} & \small{56.98} & \small{76.75} & \small{74.33} & \small{57.25} & \small{78.40} & \small{8.91} & \small{9.70} & \small{7.15} & \small{9.03} & \small{9.86} & \small{7.34}\\\hline
\small{Utiyama} & \small{All Files} & \small{70.71} & \small{54.62} & \small{76.11} & \small{65.84} & \small{47.67} & \small{71.27} & \small{14.58} & \small{17.08} & \small{11.53} & \small{14.95} & \small{17.53} & \small{11.95}\\\hline
\small{Kehagias } & \small{Set*1(3-11)} & \small{64.90} & \small{53.88} & \small{70.12} & \small{61.77} & \small{51.04} & \small{67.92} & \small{15.69} & \small{15.05} & \small{13.12} & \small{17.16} & \small{17.208} & \small{14.67}\\\hline
\small{Kehagias} & \small{Set*2(3-5)} & \small{85.13} & \small{62.94} & \small{87.58} & \small{85.11} & \small{49.15} & \small{87.48} & \small{6.45} & \small{9.16} & \small{5.15} & \small{6.52} & \small{13.68} & \small{5.24}\\\hline
\small{Kehagias} & \small{Set*3(6-8)} & \small{90.51} & \small{80.51} & \small{92.29} & \small{90.51} & \small{80.51} & \small{92.29} & \small{2.54} & \small{2.76} & \small{2.04} & \small{2.47} & \small{2.69} & \small{1.96}\\\hline
\small{Kehagias} & \small{Set*4(9-11)} & \small{91.92} & \small{83.88} & \small{93.11} & \small{91.92} & \small{83.88} & \small{93.11} & \small{1.29} & \small{1.59} & \small{1.10} & \small{1.23} & \small{1.534} & \small{1.03}\\\hline
\small{Kehagias} & \small{All Files} & \small{83.12} & \small{70.30} & \small{85.78} & \small{82.33} & \small{66.15} & \small{85.2} & \small{6.49} & \small{7.14} & \small{5.35} & \small{6.84} & \small{8.78} & \small{5.73}\\\hline
\small{Affinity Prop} & \small{Set*1(3-11)} & \small{24.34} & \small{15.43} & \small{24.5} & \small{17.72} & \small{11.8} & \small{17.85} & \small{-} & \small{-} & \small{-} & \small{33.31} & \small{34.08} & \small{33.29}\\\hline
\small{Affinity Prop} & \small{Set*2(3-5)} & \small{27.41} & \small{25.53} & \small{27.41} & \small{24.67} & \small{23.44} & \small{24.67} & \small{-} & \small{-} & \small{-} & \small{N/A} & \small{5.80} & \small{N/A}\\\hline
\small{Affinity Prop} & \small{Set*3(6-8)} & \small{22.70} & \small{29.47} & \small{22.71} & \small{22.33} & \small{29} & \small{22.33} & \small{-} & \small{-} & \small{-} & \small{31.12} & \small{31.11} & \small{31.07}\\\hline
\small{Affinity Prop} & \small{Set*4(9-11)} & \small{15.43} & \small{20.47} & \small{15.48} & \small{15.42} & \small{20.4} & \small{15.48} & \small{-} & \small{-} & \small{-} & \small{25.55} & \small{28.01} & \small{25.42}\\\hline
\small{Affinity Prop} & \small{All Files} & \small{22.47} & \small{15.43} & \small{22.53} & \small{20.03} & \small{11.8} & \small{20.08} & \small{-} & \small{-} & \small{-} & \small{22.5} & \small{34.08} & \small{22.44}\\\hline
\end{tabulary}
\linebreak
\textbf{Table 12}:Precision, Recall, Beeferman's PK and WindowDiff values (per cent) obtained by the four algorithms in the first group of experiments, without and with use of named entities for Greek texts as well as use of Lucarelli's NE annotation tool only. 
\end{landscape}
\pagebreak

\subsubsection{ Second Group of Experiments}

In order to test the validity of our expectation in a more complex task, we created a second collection, which also uses as source the Stamatatos corpus (\cite{Stamatatos:2001}). In this collection, a single dataset is constructed which contains 200 texts, where every author is represented in each text. Each text is the concatenation of ten segments. More specifically, the construction of each segment is performed as follows:
\begin{itemize}
\item  An author among the ten, named I is randomly selected.
\item  A text (named k) among the thirty available belonging to the I author, is randomly selected. Let Z be the number of paragraphs that the k-th text contains.
\item  A number l (1<l<Z), corresponding to the number of paragraphs that the generated segment will contain, is randomly selected.
\item  A number m (1<m<Z-l), corresponding to the 'starting paragraph' was randomly selected. Thus, the segment contains all the paragraphs of text k starting from paragraph m and ending at the paragraph m + l.
\end{itemize}

The procedure described above produced segments and consequently concatenated texts that were longer than the ones used in the first group of experiments. Hence, the segmentation task here is more difficult. Table 13 lists the values of Precision, Recall,  Beeferman's \textit{PK} and WindowDiff metrics reported in the literature, after applying Choi’s  C99b, Utiyama \& Isahara's, Kehagias et al. and Affinity Propagation algorithms on the original (i.e., non-annotated) corpus, the output produced after applying Lucarelli's annotation tool only, as well as the respective values after applying the same algorithms on this unique dataset, where annotation including co-reference resolution was previously performed. Italic and underlined notation is used for obtained results for all files and all cases in every segmentation algorithm.

Table 13 reveals that segmentation performance was improved in the annotated corpus using both NE and co-reference resolution for all accuracy metrics and for all algorithms, with the exception of Window Diff performance for the Utiyama \& Isahara's algorithm, where a slight decrease is observed. An explanation is that in this dataset, the segment length is high leading to a high number of named entity instances. For all metrics and for all algorithms, increase in performance is achieved using co-reference resolution. 

Special attention must be given to co-reference resolution, which proved to have a significant contribution in the increase of the number of named entity instances per segment. Experiments using only Lucarelli's et al. annotation tool exhibit lower performance compared with the non-annotated corpus, except for all values of Beeferman's \textit{Pk} as well as WindowDiff performance obtained by Choi’s C99b and Affinity Propagation algorithms. This can be attributed to the fact that, annotation tools augment text vocabulary with the presence of unnecessary tags. Additionally, obtained performance by the application of co-reference resolution step at the output produced by Lucarelli's et al. NER tool, proves the effectiveness and importance of both steps. This results from obtained results in all metrics and for all algorithms, except for Affinity Propagation's WindowDiff performance, where a slight decrease is observed.

The type of named entity instance for both groups of experiments acts once again
indirectly as a discriminative factor in the segmentation process. Since, as we mentioned earlier for English, the type thus the number of occurrences of named entity instances is strongly related to document's topic, documents belonging for example to Kiosse author - whose texts are about archaeology - have more named entity instances of person name, date, and location than of group name. The opposite holds for Alachiotis documents which are about biology, where group name is the most frequent named entity type. Co- reference resolution proved once again to enhance segmentation accuracy by reinforcing the presence and frequency of appearance of named entity instances (i.e., words and phrases) and consequently, intra-segment similarity. The obtained performance of segmentation algorithms measured by WindowDiff metric in the current dataset in both groups of experiments can be also be attributed to the annotation performed using the Lucarelli, Vasilakos and Androutsopoulos (2007) annotation tool, which was trained on similar topics. As a consequence, no correction was required since the tool was proven to perform correct annotations to those named entity instances that could identify and appropriately classify.
\pagebreak
\begin{landscape}
\begin{tabulary}{1.25\textwidth}
{|l|l|l|l|l|l|l|l|l|l|l|l|l|}\\\hline
\small{\textbf{Algorithm}} & \small{\textbf{Precision}} & \small{\textbf{Precision}} & \small{\textbf{Precision}} &\small{\textbf{Recall}} & \small{\textbf{Recall}} & \small{\textbf{Recall}} &\small{\textbf{Pk}} & \small{\textbf{Pk}} & \small{\textbf{Pk}}&\small{\textbf{WindowDiff}} & \small{\textbf{WindowDiff}} & \small{\textbf{WindowDiff}}\\\hline
\textbf{} &\small \textbf{no} &\small \textbf{NES} & \small\textbf{with} &\small\textbf{no} & \small\textbf{NES} & \small\textbf{with} &\small\textbf{no} & \small\textbf{NES} & \small\textbf{with}&\small\textbf{no} & \small\textbf{NES} & \small\textbf{with}\\\hline
\textbf{} & \small\textbf{annotat} & \small\textbf{annotat} & \small\textbf{annotat} &\small\textbf{annotat} & \small\textbf{annotat} & \small\textbf{annotat} &\small\textbf{annotat} & \small\textbf{annotat} & \small\textbf{annotat}&\small\textbf{annotat} & \small\textbf{annotat} & \small\textbf{annotat}\\\hline
\small{Choi's C99b} & \small{44.62} & \small{39.65} & \small{49.40} & \small{44.62}& \small{39.65} & \small{49.40} & \small{19.44} & \small{19} & \small{18.12} & \small{21.62} & \small{21} & \small{20.47}\\\hline
\small{Utiyama} & \small{56.76} & \small{47.15} & \small{59.78} & \small{67.22}& \small{55.05} & \small{69.00} & \small{12.28} & \small{11.84} & \small{10.83} & \small{12.26} & \small{14.77} & \small{13.57}\\\hline
\small{Kehagias} & \small{60.60} & \small{49.90} & \small{63.46} & \small{57.00}& \small{48} & \small{62.00} & \small{11.07} & \small{11} & \small{9.06} & \small{11.06} & \small{13.40} & \small{9.30}\\\hline
\small{Affinity Prop} & \small{8.83} & \small{2.82} & \small{9} & \small{14.64}& \small{5} & \small{14.91} & \small{-} & \small{-} & \small{-} & \small{57.38} & \small{57.22} & \small{57.30}\\\hline
\end{tabulary}
\linebreak

\textbf{Table 13}:Precision, Recall, Beeferman's PK and WindowDiff values (per cent) obtained by the four algorithms in the second group of experiments, without and with use of named entities for Greek texts as well as use of Lucarelli's NE annotation tool only. 
\end{landscape}
\pagebreak

\section{Conclusions - Future Work}

In this paper, we evaluated the benefit of incorporating information extraction techniques to enhance the performance of text segmentation algorithms. More specifically, we performed manual NER and co-reference resolution on an English corpus (Choi's benchmark) used by text segmentation algorithms. We also examined two combinations of publicly available automated annotation tools. The first combination involves the output of Illinois NER and Reconcile Co-referencer, while the second involves the output of Illinois NER and Illinois Co-referencer. The output produced by the aforementioned combinations as well as the one produced by applying Illinois NER only, is appropriately post-processed by a dedicated manually contructed parser by the author, in order to be given as input to the segmentation algorithms. For English, a general conclusion is that all segmentation algorithms work better with manual annotation, (compared to original raw texts). Moreover, segmentation results obtained with automatic annotations strongly depend on the tools used and are sometimes worse than using just the raw texts.

For Greek, we performed manual completion regarding NER and mainly co- reference resolution after applying an automated NE annotation tool on a manually constructed corpus. We then compared the performance of four well-known text segmentation algorithms in both the original and the resulting annotated corpora (using only named entity annotation and combination of named entity annotation and co- reference resolution). The results obtained for this case show that, combination of named entity annotation and co-reference resolution has an added value as the segment length increases.

The degree to which automatic NER and co-reference system may help is strongly related to the following factors: a) the nature of the dataset for segmentation used; b) the NER system, which ideally should be trained on similar corpus; c) the number of named entity types used. An intuition is that, use of an important number of “quality” named entity types may affect named entity (instance) distribution thus, statistical (and semantic) coherence of a segment; d) the types of co-reference that a co-reference system captures; e) the combination of automated tools used (assumptions (a) to (e)) are reinforced by statements presented by Atdağ and Labatut (\cite{Atdag:2013}), Appelt (\cite{Appelt:1999}), Siefkes (\cite{Siefkes:2007}) and Marrero et al. (\cite{Marrero:2009})); f) the validity of the constructed parser to transform a "tagged" output resulted from the application of a NER and a co-reference system to plain text, which is necessary in order to be given as input to a text segmentation algorithm.

On the other hand, in the Greek corpus, the obtained performance of four well known text segmentation algorithms exhibit a high benefit when annotation is exploited. This is apparent in all algorithms and for all metrics and datasets which can be attributed to the appropriateness of the automated annotation tool applied. The contribution of co- reference resolution in this improvement is high and deserves special attention. The latter has an added value in languages such as Greek, which is a high inflectional language. The benefit of performing manual annotation instead of automatic annotation was also examined. It must be stressed that, automatic NER including co-reference resolution can be performed for Greek as long as a good combination of tools is accomplished. However, to the author’s best knowledge, for Greek only Papageorgiou et al. (\cite{Papageorgiou:2002}) work performing co-reference resolution appears in the literature. Since NER is strongly related to corpus topic(s), use of publicly available automated annotation tools may lead to unexpected results when the automated annotation tool is applied in a corpus involving different topics than the one(s) the tool was trained to.

Obtained results in all experiments for both languages prove that the benefit of named entity annotation and co-reference resolution is apparent since it reinforces named entity instances and mentions of them because: (a) it highlights expressions attributed to named entity types which otherwise are treated as separated words; (b) it preserves words such as pronouns which otherwise are eliminated as a result of stop list removal and stemming; (c) it augments named entity instance frequency and consequently distribution, thus intra segment similarity as a result of co-reference resolution; (d) it makes named entity types to act indirectly as a discriminative factor in the segmentation process, since words are categorized to the most appropriate named entity type.

Moreover, named entity annotation and co-reference resolution are strongly related to the segment's topic as well as the number of named entities instances appearing in it. Since current experiments consists a preliminary study, we cannot draw a firm conclusion regarding statistical significance of the obtained results. For some portions of the datasets and for some algorithms statistical significance is observed while for others is not. Small improvements can be highly related to the validity of the constructed parsers. For algorithms such as Kehagias it seems to worth the effort for both English and Greek datasets while for others (such as Affinity propagation) probably not.

The general conclusion of the work presented in this paper is that, NER and co- reference resolution provide as output valuable information for the text segmentation task. The issue here is how to obtain this valuable information (especially for English). Manual annotation seems to be a more effective solution. On the other hand, a firm conclusion cannot be derived when using publicly available annotation tools (for English), whose outcome might be problematic for several reasons. The question that arises is whether the burden of choosing the appropriate annotation tools and constructing a dedicated parser to process the output produced by those tools is comparable to performing manual annotation. A possible solution to this is the examination of bootstrapping methods/algorithms in order to adapt already existing annotation tools using a small number of training data in cases where full trained ones are not effective for the task in question.

The approach followed in this paper may be beneficial for other problems, such as noun- phrase chunking, tutorial dialogue segmentation, social media segmentation (e.g., Twitter or Facebook posts), text summarization, semantic segmentation, web content mining, information retrieval, speech recognition, and focused crawling.

We consider several directions of future work. The first direction regarding English, considers performing text segmentation on a different corpus with fewer topics, such as the Reuters RCV1 and RCV2 corpora, as well as the ones compiled by Malioutov and Barzilay (\cite{Malioutov:2006}) which consists of manually transcribed and segmented lectures on Artificial Intelligence (\url{http://www.mit.edu/~igorm/thesis.pdf}), Eisenstein and Barzilay (\cite{Eisenstein:2008}) which consists of 227 chapters from medical textbooks and those originated from Project Gutenberg in which the segment boundaries correspond to chapter breaks or to breaks between individual stories. Use of other dataset(s) is considered as long as an automated annotation tool(s) trained on a similar corpus with them is used, or an already annotated corpus exists. 

The same direction can be followed for Greek i.e., performing text segmentation in the corpus used by Lucarelli, Vasilakos and Androutsopoulos (\cite{Lucarelli:2007}) containing fewer topics and in the one used by Papageorgiou et al. (\cite{Papageorgiou:2002}), where co- reference resolution was also performed. 

The second direction for both languages is oriented towards the examination of other NER systems with special attention to those containing co-reference resolution tools especially for English, where more tools are publicly available. More specifically, the author’s intention is to find an efficient combination of tools performing NER and co- reference resolution whose output can be easily be given as input to a text segmentation algorithm (without ideally the need of a parser). Segmentation accuracy can prove to act as a criterion for the selection of those tools. Regarding co-reference resolution, specific focus will be given to the types of co-reference examined as well as their scope (i.e., the examination of co-reference within the same sentence and/or with the previous one appearing in the text) as well as the contribution of each type of anaphora separately. Regarding the contribution of each type of co-reference resolution, segmentation accuracy improvement can also act as a selection criterion. The ultimate goal is to examine the benefit of applying automatic co-reference annotation tools in real-world applications.

A third direction involves the examination of bootstrapping methods/algorithms in order to adapt already existing annotation tools using a small number of training data.
Finally, we consider examining the addition of other types of named entities that will be more oriented to the segment's topic. In the same direction lies the extraction and annotation of relations between named entities and the examination of their contribution in the segmentation process in align with the work presented in Singh et al. (\cite{Singh:2013}). Our aim is to reinforce the role and identity of named entities in the segmentation process.

\bibliographystyle{alpha}
\bibliography{main}

\end{document}